\documentclass{article}

\PassOptionsToPackage{numbers, compress}{natbib}

\usepackage[final]{neurips_2025}

\usepackage[utf8]{inputenc} 
\usepackage[T1]{fontenc}    
\usepackage{hyperref}       
\usepackage{url}            
\usepackage{booktabs}       
\usepackage{amsfonts}       
\usepackage{nicefrac}       
\usepackage{microtype}      
\usepackage{xcolor}         

\usepackage{color, colortbl}
\definecolor{LightCyan}{rgb}{0.88,1,1}
\definecolor{LightGreen}{rgb}{0.56, 0.93, 0.56}
\newcommand{\CC}[1]{\cellcolor{LightGreen}}
\usepackage{amsmath}
\usepackage{enumitem}
\usepackage{relsize} 

\usepackage{titletoc}

\usepackage{amsmath}
\usepackage{amssymb}
\usepackage{mathtools}
\usepackage{amsthm}
\usepackage{multirow}
\usepackage{graphicx}
\usepackage{wrapfig}
\usepackage{tablefootnote}

\usepackage{algorithm}
\usepackage[noend]{algpseudocode}
\usepackage{amsmath} 

\usepackage[capitalize,noabbrev]{cleveref}

\theoremstyle{plain}
\newtheorem{theorem}{Theorem}[section]

\theoremstyle{definition}

\theoremstyle{remark}

\usepackage{todonotes}

\title{Looking Beyond the Known: Towards a Data Discovery Guided Open-World Object Detection}

\author{%
  Anay Majee \qquad \qquad
  Amitesh Gangrade\thanks{Work done as a graduate student at UTDallas.} \qquad \qquad
  Rishabh Iyer \\
  The University of Texas at Dallas\\
  \texttt{firstname.lastname@utdallas.edu} \\
}

\begin{document}

\maketitle

\begin{abstract}
    Open-World Object Detection (OWOD) enriches traditional object detectors by enabling continual discovery and integration of unknown objects via human guidance. 
However, existing OWOD approaches frequently suffer from semantic confusion between known and unknown classes, alongside catastrophic forgetting, leading to diminished unknown recall and degraded known-class accuracy. 
To overcome these challenges, we propose \textit{\textbf{C}ombinato\textbf{r}ial \textbf{O}pen-\textbf{W}orld \textbf{D}etection} (\textbf{CROWD}\footnote{Project Page at \url{https://anaymajee.me/assets/project_pages/crowd.html}}), a unified framework reformulating unknown object discovery and adaptation as an interwoven combinatorial (set-based) data-discovery (CROWD-Discover) and representation learning (CROWD-Learn) task. 
CROWD-Discover strategically mines unknown instances by maximizing Submodular Conditional Gain (SCG) functions, selecting representative examples distinctly dissimilar from known objects. 
Subsequently, CROWD-Learn employs novel combinatorial objectives that jointly disentangle known and unknown representations while maintaining discriminative coherence among known classes, thus mitigating confusion and forgetting. 
Extensive evaluations on OWOD benchmarks illustrate that CROWD achieves improvements of 2.83\% and 2.05\% in known-class accuracy on M-OWODB and S-OWODB, respectively, and nearly 2.4$\times$ unknown recall compared to leading baselines.\looseness-1

\begin{figure*}[hb]
        \centering
        \includegraphics[width=\textwidth]{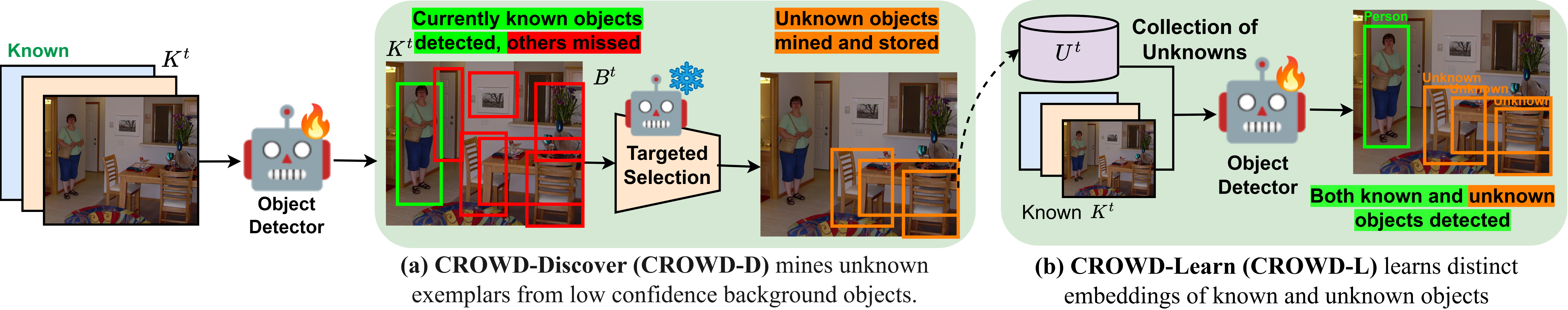}
        \vspace{-3ex}
        \caption{\small \textbf{Overall Architecture of CROWD} showing our novel combinatorial data-discovery guided representation learning approach to (a) identify unknown objects$^3$ and (b) learn distinguishable representations of both known and unknown objects.\looseness-1}
        \label{fig:spotlight_fig}
        \vspace{-5ex}
\end{figure*}
\end{abstract}

\section{Introduction}
Object Detection (OD) is central to numerous vision applications~\cite{clinical, majee2021fewshot, koh2020wilds}, but as shown in \cref{fig:spotlight_fig} (left) conventional OD models operate under a closed-world assumption, where the object vocabulary remains fixed throughout deployment. This limitation hinders the model’s ability to generalize to novel object categories. \textit{Open-World Object Detection} (OWOD), introduced by~\citet{joseph2021ore}, addresses this by combining open-set recognition~\cite{wang2024dissect,zhang2024learning,wang2024os} with incremental learning~\cite{li2025adaptive,fukuda2024adapter}, enabling models to detect unknown objects and subsequently recognize them with minimal supervision, thus supporting continual self-improvement. However, recent efforts~\cite{Randbox, Zohar_2023_CVPR, sun2024exploring} reveal two enduring challenges: (1) confusion between known and unknown objects~\cite{joseph2021ore, gupta2021ow, cat}, and (2) catastrophic forgetting of previously learned classes~\cite{sun2024exploring, Randbox, allow}. Confusion arises due to visual similarity between unknown and known classes (e.g., truck vs. car headlamps), while forgetting stems from the lack of supervision for unknowns, causing them to be misclassified as background. Existing methods struggle to address both issues, as evidenced by low unknown recall and high Wilderness Impact scores~\cite{joseph2021ore} (\cref{tab:crowd_confusions}). These limitations motivate the need for a framework that can effectively discover unknown Region-of-Interests (RoIs) and learn representations that remain distinct from known categories, thereby mitigating confusion and preserving prior knowledge.\looseness-1


We cast Open-World Object Detection (OWOD) as a set-based discovery and learning problem. For each task $t$, we view the \textit{known} object classes as a collection of sets $K^t$, group all candidate \textit{unknowns} into a single pseudo-labeled set $U^t$~\cite{joseph2021ore}, and treat everything else as \textit{background} $B^t$. 
This formulation (\cref{sec:prob_def}) facilitates the incorporation of submodular functions into OWOD and gives rise to our \textbf{C}ombinato\textbf{r}ial \textbf{O}pen-\textbf{W}orld \textbf{D}etection (\textbf{CROWD}) framework. As illustrated in \cref{fig:spotlight_fig}, CROWD tackles OWOD as an interleaved process of data discovery (CROWD-D) and representation learning (CROWD-L), directly targeting the dual challenges of confusion and forgetting.\looseness-1

Starting from a OD trained only on known instances $K^t$ (\cref{fig:spotlight_fig}\footnote{\cref{fig:spotlight_fig}(a) shows a subset of background and unknown RoIs for clarity. The total number of RoIs in the original experiment is set to 512 (as in \cite{sun2024exploring}) while the number of mined unknowns is set to 10 (per image).}), CROWD-Discover (CROWD-D) identifies representative unknowns, formulated as a combinatorial targeted selection problem. Specifically, CROWD-D maximizes the Submodular Conditional Gain (SCG, \cref{sec:submod_prelims}) between $K^t$ and candidate subsets, encouraging dissimilarity with known and background objects. Authors in \cite{prism} provide theoretical guarantees that greedy maximization~\cite{Mirzasoleiman2015lazierthanlazy} of SCG results in selection of samples in $U^t$ which are \textit{dissimilar} to $K^t$ as well as $B^t$, constituting informative pseudo-labeled unknowns.\looseness-1

CROWD-Learn (CROWD-L) then fine-tunes the OD model on both known and mined unknowns as shown in \cref{fig:spotlight_fig}(b) via a novel combinatorial joint objective (\cref{eq:crowd_loss}), rooted in two families of submodular functions: SCG~\cite{prism} and Total Submodular Information~\cite{fujishige} (SIM). Maximizing SCG increases diversity between known and unknown objects, reducing feature overlap and confusion, as corroborated by \cref{tab:owod_ablations}. Conversely, minimizing SIM encourages intra-class cohesion within each known object, preserving discriminative features and mitigating forgetting. 
This formulation closely follows the observation in \cite{score, smile} that submodular functions model cooperation~\cite{submod_cooperation} and diversity~\cite{submod_diversity} when minimized and maximized respectively.
Finally, we instantiate a family of submodular-based loss functions within CROWD-L that jointly reduce confusion and forgetting, achieving notable gains in unknown recall and known-class accuracy (\cref{tab:owod_ablations}). We validate our approach on two standard OWOD benchmarks, M-OWOD~\cite{joseph2021ore} and S-OWOD~\cite{gupta2021ow}, demonstrating its effectiveness across diverse open-world settings. Our main contributions are -\looseness-1

\begin{itemize}[leftmargin=*]
    \item  CROWD introduces a novel combinatorial viewpoint in OWOD by \textbf{modeling the identification of unknown instances of a given task as a data discovery problem} (CROWD-D), selecting unknown RoIs which maximize the SCG between and the known object instances.\looseness-1
    \item CROWD also introduces a novel \textbf{set-based learning paradigm CROWD-L, based on SCG functions which minimizes the cluster overlap between embeddings of known and unknown objects} while retaining the discriminative feature information from the known ones.\looseness-1
    \item Finally, CROWD demonstrates $\sim$2.4$\times$ increase in unknown recall per task alongside up to 2.8\% improvement on M-OWODB and 2.1\% improvement on S-OWODB in known class performance (measured as mAP) over several existing OWOD baselines.\looseness-1
\end{itemize}

\section{Related Work}
\noindent \textbf{Open-World Object Detection} (OWOD) first introduced in \citet{joseph2021ore} augmented a Faster R-CNN~\cite{frcnn} model with contrastive clustering and an Energy-Based Unknown Classifier relying on a objectness threshold based pseudo labeling strategy. Subsequent work such as OW-DETR~\cite{gupta2021ow} adapted deformable DETR~\cite{zhu2020deformable} and proposed an attention-based pseudo-labeling scheme that identifies high-activation regions as unknowns without requiring extra supervision. 
Further, CAT~\cite{Ma2023CATLA} improves transformer-based models by decoupling localization and classification, while introducing dual pseudo-labeling strategies, namely - model-driven and input-driven—to robustly mine unknowns. 
PROB~\cite{Zohar_2023_CVPR} advanced the state of the art by modeling objectness probabilistically in the embedding space using a Gaussian likelihood, allowing better separation of unknowns from background without explicit negative examples. 
Other notable works include 2B-OCD~\cite{wu2022two}, which integrates a localization-based objectness head~\cite{kim2021openproposals}, OCPL~\cite{yu2022open} enforces class-prototype separation to reduce known-unknown confusion, and UC-OWOD~\cite{wu2022ucowod} employs feature-space regularization to suppress background misclassification. 
Some recent methods leverage external supervision, e.g., MViTs~\cite{maaz2022multimodal}, or multimodal cues such as text for class-agnostic detection, these often fall outside strict OWOD assumptions but highlight promising directions for future research. 
Complementary to these, recent approaches such as RandBox~\cite{Randbox} sidesteps detection bias via random bounding box sampling and dynamic-k filtering, while OrthogonalDet~\cite{sun2024exploring} enforces angular decorrelation in object features to disentangle objectness and class semantics.
Interestingly, Randbox and OrthogonalDet outperforms larger models like OW-DETR, UC-OWOD etc. while using a simpler Faster-RCNN~\cite{frcnn} based architecture.
Despite substantial progress, OWOD methods continue to grapple with confusion between known and unknown objects and catastrophic forgetting during incremental adaptation recently evidenced in \citet{xi2024umb}, motivating the development of our CROWD framework.\looseness-1
In general our work is also related to standard object detection while CROWD-Discover (\cref{sec:crowd_d}) is related to combinatorial subset selection, the related work for which is provided in \cref{app:additional_rel_work}.\looseness-1

\section{Method}
\subsection{Problem Definition: OWOD}
\label{sec:prob_def}
\noindent We largely adopt the problem formulation of OWOD from \citet{joseph2021ore} with modifications towards a combinatorial (set-based) formulation. Given an incoming task $T_t$ where $t \in [1, n]$, an object detector $h^t(.; \theta)$ recognizes a set of known classes $K^t = \{K^t_1, K^t_2, \dots, K^t_{C^k}\}$, $|K^t| = C^k$ while also accounting for unknown classes $U^t$ that may appear during inference (classes in $U^t$ are not labeled during training). Here, $K^t_i$, $i \in [1, C^k]$ indicates examples for each known class in $T_t$. 
The dataset $D^t = \{(x^t_i, y^t_i)\}_{i = 1}^{M}$ for each task $T_t$, where each label $y^t_i$ contains $K$ object instances ($K$ can very for each image) defined by bounding box parameters 
$y^t_k = [c_k, x_k, y_k, w_k, h_k]$, with $c_k \in [1, C^k]$ representing the class label. 
The object detection model $h^t(.; \theta)$ is trained to learn newly introduced instances from labeled examples in $K^t$ while identifying unknown objects $U^t$ by assigning them a placeholder label ($0$). 
Examples in $U^t$ can be reviewed by a human expert who identifies $C^u$ new classes, allowing the model to update incrementally and produce $h^{t + 1}(.; \theta)$ without retraining on the entire dataset. If $\hat{U^t}$ indicate the newly labeled set of unknown classes s.t. $|\hat{U^t}| = C^u$, then $K^{t+1} = K^t \cup \hat{U^t}$, enabling continual adaptation to new object categories over time.\looseness-1

\subsection{Preliminaries: Submodularity}
\label{sec:submod_prelims}
\noindent Adopting a set-based formulation allows us to explore combinatorial functions for OWOD. 
In particular we explore \textit{Submodular functions} which are set functions exhibiting a unique diminishing returns property. Formally, a function $f: 2^\mathcal{V} \rightarrow \mathbb{R}$ defined on a ground set \(\mathcal{V}\) is submodular if for any subset $A_i, A_j \subseteq \mathcal{V}$, it holds that $f(A_i) + f(A_j) \geq f(A_i \cup A_j) + f(A_i \cap A_j)$ \citep{fujishige}.
These functions have been widely studied for applications such as data subset selection~\citep{killamsetty_automata, prism, jain2023efficient}, active learning~\citep{wei15_subset, talisman, Beck2021EffectiveEO, Kaushal_2019}, and video summarization~\citep{vid_sum_2019, kaushal2019framework}.
Typically, these tasks involve formulating subset selection or summarization as submodular maximization~\citep{fujishige, Nemhauser1978} subject to a knapsack constraint~\citep{Nemhauser1978}. A classic result guarantees a \((1 - e^{-1})\) approximation factor~\citep{Nemhauser1978} using a greedy algorithm, which can be implemented more efficiently via improved greedy strategies~\citep{Mirzasoleiman2015lazierthanlazy}.
Within this framework, \textbf{Submodular Information Functions} (SIMs)~\citep{fujishige}, such as Facility-Location or Graph-Cut, promote diversity when maximizing \(f(A)\). 
On the other hand, \textbf{Submodular Conditional Gain} (SCG), \(H_f(A_i \mid A_j)\), captures elements in \(A_i\) most dissimilar to \(A_j\). 
Extrapolating this, \citet{kothawade2022add} defines discovery of unseen, rare examples as a targeted selection problem. 
Further works~\cite{score, smile} have demonstrated the utility of these combinatorial functions in continuous optimization. \citet{score} introduces losses inspired by SIMs to enforce intra-group compactness (when minimized) and inter-group separation (when maximized), while \cite{smile} uses SMI-based losses to account for interactions between abundant and rare samples in few-shot learning.
\textit{Motivated by these insights, CROWD proposes a novel data-discovery guided representation learning framework to dynamically identify and incrementally adapt to unknown objects}.\looseness-1

\subsection{The CROWD Framework}
The problem formulation in \cref{sec:prob_def} surfaces two unique challenges in the domain of OWOD - \textbf{(1)} How to \textit{identify instances of unlabeled unknown objects} $U^t$ given labeled examples of only known ones in $K^t$ ? \textbf{(2)} How to effectively learn representations of currently known objects \textit{without forgetting the previously known} (classes introduced in $T_i$, where $i < t$) ones ?
\begin{wrapfigure}{r}{0.6\textwidth}
\vspace{-3ex}
\begin{minipage}{0.6\textwidth}
    \begin{algorithm}[H]
    \caption{Discovering Unknown RoIs in CROWD-D} \label{alg:crowd_discover}
    \begin{algorithmic}[1]
    \Require A task $t$, set of RoI feature vectors $\mathtt{R} \in \mathbb{R}^{N \times d}$, Objectness scores $\mathbb{S}(.) \in \mathbb{R}^{N}$, Task specific Labels $y_{K^t}$, budget $\mathtt{k}$
    
    \State \textcolor{blue}{\textit{/** Identify and Exclude outliers **/}} 
    \State $\mathtt{R} \leftarrow \{r \in \mathtt{R} | \mathbb{S}(r) \geq \tau_e\}$
    
    \State $K^t \leftarrow \texttt{HUNGARIAN-MATCHING}(\mathtt{R}, y_{K^t})$ \Comment{\textcolor{gray}{\textit{Known class RoI features}}}
    \State $\mathcal{V} \leftarrow \mathtt{R} \setminus K^t$

    \State \textcolor{blue}{\textit{/** Select Background Samples **/}} 
    \State $B^t \leftarrow \underset{\substack{B^t \subseteq \mathcal{V} \\ |B^t| \leq \tau_b\%|\mathcal{V}|}}{\arg\max}~ H_f(B^t \mid K^t)$ \Comment{\textcolor{gray}{\textit{Large feature separation from $K^t$}}}

    \State \textcolor{blue}{\textit{/** Select Unknown samples from $\mathtt{R} \setminus K^t$ **/}} 
    \State $U^t \leftarrow \underset{U^t \subseteq \mathcal{V}, |U^t| \leq \mathtt{k}}{\arg\max}~ H_f(U^t \mid K^t \cup B^t)$ \Comment{\textcolor{gray}{\textit{Unknowns are different from $K^t \cup B^t$}}}
    
    \State \Return $U^t$
    
    \end{algorithmic}
    \end{algorithm}
\end{minipage}
\vspace{-2ex}
\end{wrapfigure}

\begin{figure*}[t]
        \centering
        \includegraphics[width=\columnwidth]{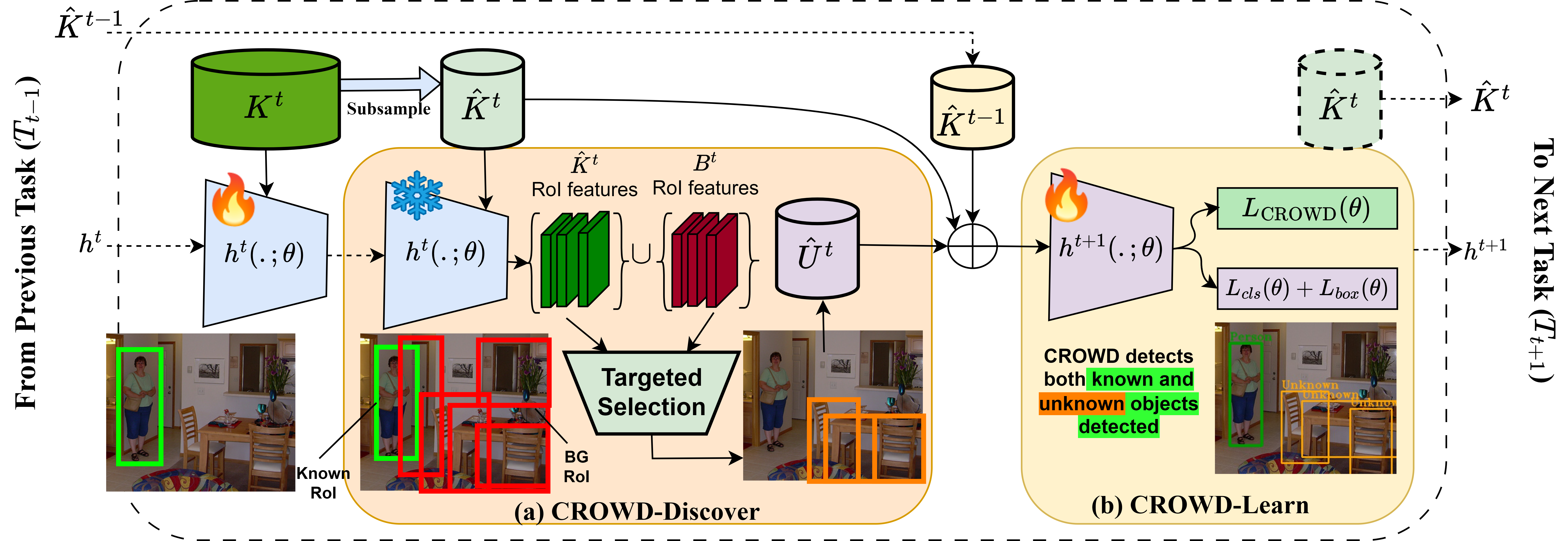}
        \vspace{-3ex}
        \caption{\small \textbf{Interleaved Data-Discovery and Representation Learning in CROWD} on an incoming task $T_t$. CROWD takes as input the model weights from $T_{t-1}$ and a small replay buffer of previously known classes $\hat{K}^{t -1}$, applies (a) CROWD-Learn to discover unknown RoIs and (b) CROWD-L to learn discriminative features of both known and unknown instances to return an updated model $h^{t+1}$ and the current task replay buffer $\hat{K}^t$.}
        \label{fig:crowd_interleaved}
        \vspace{-2ex}
\end{figure*}

To this end, we introduce \textbf{C}ombinato\textbf{r}ial \textbf{O}pen-\textbf{W}orld \textbf{D}etection (\textbf{CROWD}) framework, which models the OWOD task as a interleaved set-based data discovery~\cite{kothawade2022add} and representation learning~\cite{score} problem. 
CROWD achieves this in two stages - namely \textbf{CROWD-Discover} (a.k.a. CROWD-D) and \textbf{CROWD-Learn} (a.k.a. CROWD-L) as shown in \cref{fig:crowd_interleaved}. 
Given an incoming task $T_t$ we first train $h^t(.; \theta)$ on currently known classes in $D^t$. At this point, CROWD-D utilizes the frozen model weights of $h^t$ and uses a small replay buffer (typically containing examples from both previously known and currently introduced objects) to \emph{discover} highly representative proposals of unknown classes $U^t$. We elucidate this in \cref{sec:crowd_d}.
Subsequently, CROWD-L introduces a novel combinatorial learning strategy to rapidly finetune $h^t$ on this replay buffer (we adopt the predefined buffer in \citet{joseph2021ore}) to distinguish between known classes $K^t$ and unknown $U^t$ while preserving distinguishable features from the previously known classes. We discuss this in detail in \cref{sec:crowd_l}.

\subsubsection{CROWD-Discover}
\label{sec:crowd_d}
\noindent During training of $h^t(.;\theta)$, label information is available only for the currently known classes $K^t$ with no labels of the previously known $K^{t-1}$ and the unknown classes $U^t$.
CROWD-D tackles the challenge of \textbf{identifying potentially unknown instances} from Region-of-Interest (RoI) proposals produced by the Region-Proposal-Network (RPN) in $h^t(.;\theta)$. Unlike existing OWOD methods employ pseudo labeling~\cite{joseph2021ore}, feature orthogonalization~\cite{sun2024exploring} etc. rely on the objectness score (probability of an RoI proposal to contain a foreground object), whereas CROWD-D achieves this by \textbf{modeling this task as a combinatorial data discovery problem}~\cite{kothawade2022add}.
\textit{Given a set of RoI proposals $\mathtt{R}$ and a submodular function $f$ we define the data discovery task (\cref{alg:crowd_discover}) as a targeted selection problem which selects a set of unknown instances $U^t$ from $\mathcal{V} = \mathtt{R} \setminus K^t$ that maximizes the SCG $H_f$ given a query set comprising of known $K^t$ and the background $B^t$ instances (line 8 in \cref{alg:crowd_discover})}.\looseness-1

\begin{figure*}[t]
        \centering
        \includegraphics[width=\textwidth]{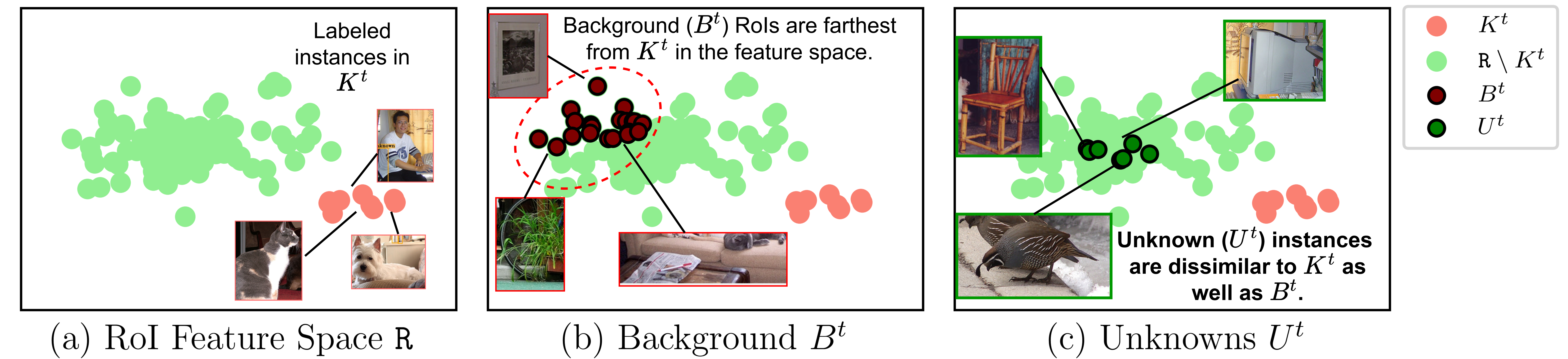}
        \vspace{-10pt}
        \caption{\small \textbf{Illustration of the data-discovery pipeline in CROWD-D} on a synthetic dataset with $|\mathtt{R}| = 500$ and budget $\mathtt{k} = 10$ and the underlying submodular function as Graph-Cut. CROWD-D selects $U^t$ which are both dissimilar to background $B^t$ and known $K^t$ instances.\looseness-1}
        \label{fig:crowd_selection_gccg}
        \vspace{-4ex}
\end{figure*}

Here, $\mathtt{k}$ denotes a budget for the number of unknown samples mined per image. 
From the definition of SCG in \cref{sec:submod_prelims} \textbf{selected examples in $U^t$ are largely dissimilar to examples in $K^t \cup B^t$} indicating that they are neither background objects nor visually similar to known objects as shown in \cref{fig:crowd_selection_gccg}(c).
\noindent Interestingly, the number of known RoIs $K^t$ in $\mathtt{R}$ are significantly fewer than the background RoIs $B^t$ (typically $|\mathtt{R}| = 500$ (total number of RoIs from the RPN) whereas $|K^t| \sim 10$ in MS-COCO~\cite{coco} which leaves $|B^t| \sim 490$) in most OD models.
To minimize the computation costs while selecting $U^t$ from this large RoI pool we exclude all RoIs with low objectness scores $\mathbb{S} < \tau_e$ (line 2 in \cref{alg:crowd_discover}) and likely background objects $B^t$ predicted with high confidence (line 6 of \cref{alg:crowd_discover}). 
Instances in $B^t$ are selected which maximize the SCG between themselves and known RoIs $K^t$ under a budget constraint of $\tau_b \% |\mathcal{V}|$ (samples that are significantly different from $K^t$).
Known instances are identified by following the hungarian matching technique applied in \citet{Randbox} as shown in line 3 of \cref{alg:crowd_discover}.
The exclusion thresholds $\tau_e$ and $\tau_b$ are empirically determined to be 0.2 and 30\% respectively and the underlying submodular function in our experiments is chosen to be Graph-Cut which has been evidenced in \citet{prism} to model both representation and diversity among selected examples.\looseness-1

\subsubsection{CROWD-Learn}
\label{sec:crowd_l}
Including unknown examples $U^t$ can potentially inject noisy labels into the training data with detrimental effects.
We show in \cref{tab:owod_ablations} that CROWD-D alone does not handle the knowledge retention from previously known classes $K^{t-1}$, despite significant improvements on unknown class recall, causing forgetting.
CROWD-L overcomes the aforementioned challenges by introducing a novel \textbf{Combinatorial representation learning} strategy inspired from recent developments~\cite{score, smile}, that ensures orthogonality (separation) between embeddings in $K^t$ and $U^t$ while minimizing the effect of forgetting of $\hat{K}^{t-1}$. Here, $\hat{K}^{t-1}$ denotes a replay buffer of previously known classes.
\begin{align}
\begin{split}
    L_{\text{CROWD}}^{self}(\theta) =& \sum_{i = 1}^{C^t} f(K^t_i; \theta) 
    \text{ ;  } \\
    L_{\text{CROWD}}^{cross} (\theta) =& \sum_{i = 1}^{C^t} H_f(K^t_i | U^t; \theta) = \sum_{i = 1}^{C^t} f(K^t_i \cup U^t) - f(U^t)\\
    L_{\text{CROWD}}(\theta) =& L_{\text{CROWD}}^{self}(\theta) - \eta L_{\text{CROWD}}^{cross}(\theta) \\
\end{split}
\label{eq:crowd_loss}
\end{align}
\textit{Given a set of known $K^t \cup \hat{K}^{t-1}$, unknown $U^t$ classes alongside a submodular function $f$ we define a learning objective $L_{\text{CROWD}}(\theta)$ as shown in \cref{eq:crowd_loss} which jointly minimizes the Submodular Total Information ($L_{\text{CROWD}}^{self}$) over each known class $K_i^t \in \{K^t \cup \hat{K}^{t-1}\}$ and the SCG ($H_f$ as defined in \cref{sec:submod_prelims}) between known class $K^t_i$ and the unknown set $U^t$ ($L_{\text{CROWD}}^{cross}$)}. 
Note that CROWD-L is applied during training of task $T_t$ as a finetuning step as shown in \cref{fig:crowd_interleaved}(b).\looseness-1
\vspace{-1ex}

Note that $f$ relies on the pairwise interaction between examples in a batch which we represent using cosine similarity $s_{ku}(\theta) = \frac{h^t(x_{k}, \theta)^{\text{T}} \cdot h^t(x_{u}, \theta)}{||h^t(x_{k}, \theta)|| \cdot ||h^t(x_{u}, \theta)||}$ and can be different from the one used in CROWD-D decided through ablations in \cref{sec:all_ablations}.
Our loss formulation in $L_{\text{CROWD}}$ follows the observation in \cite{score} which entails that submodular functions model cooperation~\cite{submod_cooperation} and diversity~\cite{submod_diversity} when minimized and maximized respectively.
By varying the choice of $f$ between popular submodular functions - Facility-Location (FL), Graph-Cut (GC) and Log-Determinant (LogDet) we introduce a family of loss functions summarized in \cref{tab:crowd_instances} and derivations in \cref{app:crowd_derivations}.  $L_{\text{CROWD}}$ is applied to the classification head of $h^t(.; \theta)$ model during all training stages described in Sec. \ref{sec:expts}. Our novel formulation entails some interesting properties -

\begin{table*}[t]
      \caption{\textbf{Summary of various instantiations of CROWD-L} by varying the submodular function $f$ in $L_{\text{CROWD}}^{cross}$ and $L_{\text{CROWD}}^{self}$. Here, $\mathcal{T}$ denotes a batch with instances from $K^t_i \cup U^t$.\looseness-1}
      \centering
      \resizebox{\textwidth}{!}{\begin{tabular}{l|c|c}
            \hline
            \textbf{Objective Name}  & 
            \textbf{Instances of $L_{\text{CROWD}}^{cross}$}& 
            \textbf{Instances of $L_{\text{CROWD}}^{self}$} \\
            \hline
      
            CROWD-GC &            
            $ \sum_{i = 1}^{C^t} \frac{1}{|\mathcal{T}|} [f(K_i^t; \theta) - 2 \lambda \nu \sum\limits_{k \in K_i^t, u \in U_i^t} s_{ku}(\theta)]$ &
            $\sum_{i = 1}^{C^t} \frac{1}{|K_i^t|}[\sum_{i \in K_i^t}\sum_{j \in \mathcal{T} \setminus U^t}s_{ij}(\theta) - \lambda \sum_{i, j \in K_i^t} s_{ij}(\theta)]$ \\

            & & \\ 
            
            CROWD-FL &
            $\sum_{i = 1}^{C^t} \frac{1}{|\mathcal{T}|}\sum\limits_{n \in \mathcal{T}} \max(\max\limits_{k \in K_i^t} s_{nk}(\theta)-\nu \max\limits_{u \in U^t} s_{nu}(\theta), 0)$ &
            $\sum_{i = 1}^{C^t} \frac{1}{|K^t_i|}\sum_{i \in \mathcal{T} \setminus K_i^t} \max_{j \in K_i^t} s_{ij}(\theta)$ \\   
            
            & & \\

            CROWD-LogDet &
            $\sum_{i = 1}^{C^t} \frac{1}{|\mathcal{T}|} \log\det(s_{K_i^t}(\theta) - \nu^2 s_{K_i^t, U^t}(\theta)s_{U^t}^{-1}(\theta)s_{K_i^t, U^t}(\theta)^T)$ &
            $\sum_{i = 1}^{C^t} \frac{1}{|K_i^t|}\log \det (s_{K_i^t}(\theta) + \lambda \mathbb{I}_{|K_i^t|})$ \\   
            \hline
      \end{tabular}}\\
      \label{tab:crowd_instances}
      \vspace{-2ex}
\end{table*}

\begin{figure*}[t]
        \centering
        \includegraphics[width=\textwidth]{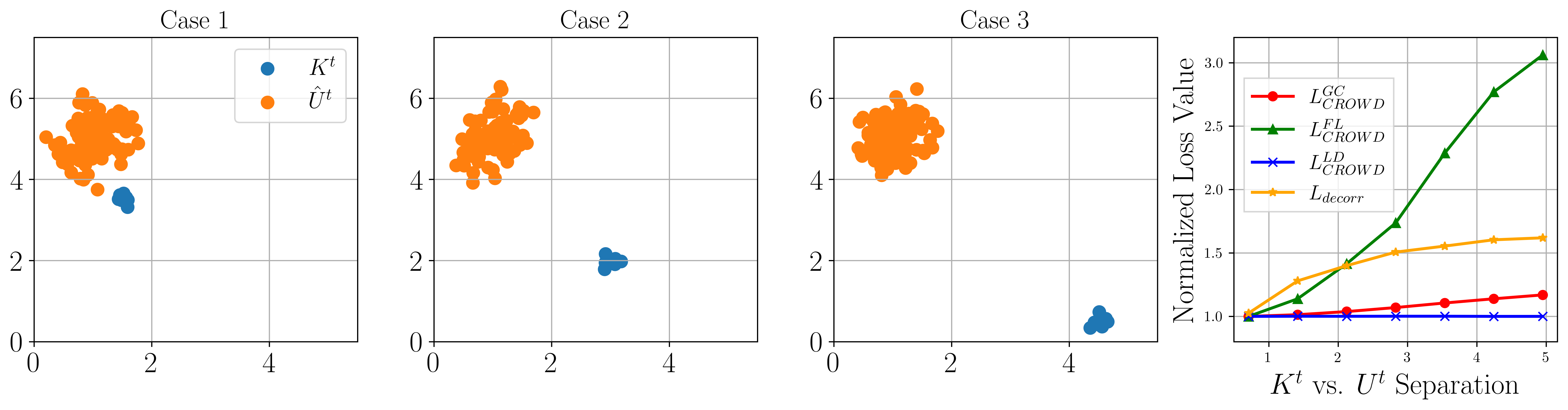}
        \vspace{-10pt}
        \caption{\textbf{Characterization of losses in CROWD-L} on a synthetic two-cluster imbalanced dataset by increasing known vs. unknown class separation (cases 1 through 3) similar to the RoI embedding space of $h^t(.; \theta)$. The synthetic dataset generation is performed under the same seed.}
        \label{fig:crowd_loss_characterization}
        \vspace{-2ex}
\end{figure*}

\noindent \textbf{(1) $L_{\text{CROWD}}^{self}$ retains informative known class features} : Following the insights in \citet{submod_cooperation} $L_{\text{CROWD}}^{self}$ which minimizes the total information contained in $K^t_i$ \textit{encourages intra-class compactness retaining the most discriminative features from the known classes alleviating forgetting}~\cite{smile}.\looseness-1

\noindent \textbf{(2) $L_{\text{CROWD}}^{cross}$ models known vs. unknown separation} : As shown in \cref{eq:crowd_loss} $L_{\text{CROWD}}^{cross}$ minimizes the SCG ($H_f$) between the embeddings of the known and unknown classes. As observed in \citet{prism} maximizing SCG models \textit{dissimilarity} between two sets. In OWOD, $L_{\text{CROWD}}^{cross}$ \textit{promotes a large inter-class boundary between $K^t_i$ and $U^t$, minimizing their cluster overlap resulting in reduced confusion}.
Further, a hyper-parameter $\eta$ controls the trade-off between intra-class compactness modeled by $L_{\text{CROWD}}^{self}$ and the inter-class separation in $L_{\text{CROWD}}$.\looseness-1

\noindent \textbf{(3) Sensitivity to unknown classes with varying $f$} : \cref{tab:crowd_instances} highlights the instances of $L_{\text{CROWD}}$ by varying $f$. Such variations injects domain specific properties into CROWD-L critical for the OWOD tasks. 
As depicted in \cref{fig:crowd_loss_characterization}, under varying known vs. unknown class separation, CROWD-FL explicitly models \textit{representation}~\cite{prism} by adopting the FL based submodular function while CROWD-LogDet injects \textit{diversity} by modeling cluster volume through Log-Determinant~\cite{alt_gen_submod}. CROWD-GC models both representation and diversity but is not resilient to imbalance between $K^t$ and $U^t$ as CROWD-FL~\cite{score}. Thus, CROWD-FL emerges to be the best choice for OWOD modeling both representation and resilience to imbalance in the embedding space over $L_{decorr}~\cite{sun2024exploring}$ (the latest baseline).\looseness-1

\noindent \textbf{(4) Generalization to Incremental Object Detection} (IOD) : As observed in several related works~\cite{joseph2021ore, gupta2021ow, sun2024exploring, Zohar_2023_CVPR}, learning from unknown pseudo labels in OWOD benefits Incremental Object Detection (IOD) tasks. However, its important to note that CROWD relies on mined unknowns (from CROWD-D) while no unknown objects are provided in the IOD setting. 
This requires us to slightly modify our learning objective $L_{\text{CROWD}}^{cross}$ to model dissimilarity between currently known $K^t$ and the previously known objects $\hat{K}^{t-1}$ (a replay buffer as in \citet{joseph2021ore}) instead of unknowns, s.t. $L_{\text{CROWD}}^{cross} (\theta) = \sum_{i = 1}^{C^t} H_f(K^t_i | \hat{K}^{t-1}; \theta)$.\looseness-1

\section{Experiments}
\label{sec:expts}
\noindent \textbf{Datasets} : We evaluate our approach on two well established benchmarks  - \textbf{M-OWOD}~\cite{joseph2021ore} and \textbf{S-OWOD}\cite{gupta2021ow}. 
M-OWOD, (\textit{Superclass-Mixed OWOD Benchmark}) consists of images from both MS-COCO~\cite{coco} and PASCAL-VOC~\cite{Everingham15} depicting 80 classes grouped into 4 tasks (20 classes per task). 
On the other hand, S-OWOD (\textit{Superclass-Separated OWOD Benchmark}) consists of images from only MS-COCO dataset. 
Both benchmarks split the underlying data points into four distinct (non-overlapping) tasks $T_t$, where $t \in [1,4]$. During training on a task $T_t$ the model in provided labeled examples from $T_t$ alone while at inference the model is expected to identify objects in tasks leading up to $T_t$, s.t $t \in [1, t]$. No prior knowledge of subsequent tasks $t \in [t+1, n]$ ($n$ refers to maximum number of tasks in an experiment) are available during training and inference on $T_t$. 
In contrast to M-OWODB, S-OWODB introduces a distinct separation between super-categories (eg. animals, vehicles etc.) and distributes these super-categories between tasks (each task will have examples from one or more unique super-categories).\looseness-1

\begin{table*}[t]
  \centering
  \caption{\textbf{Open-world object detection results across incremental tasks.} U-Recall and mAP (\%) are reported for various baselines on M-OWOD and S-OWOD benchmarks. Best results are in \textbf{bold}.}
  \resizebox{\textwidth}{!}{%
  \begin{tabular}{l|cc|cccc|cccc|ccc}
    \toprule
    \multirow{3}{*}{\textbf{Method}} & 
    \multicolumn{2}{c|}{\textbf{Task 1}} & 
    \multicolumn{4}{c|}{\textbf{Task 2}} & 
    \multicolumn{4}{c|}{\textbf{Task 3}} & 
    \multicolumn{3}{c}{\textbf{Task 4}} \\ \cline{2 - 14} 

    & \multirow{2}{*}{\textbf{U-Recall}} & \cellcolor{lightgray}\textbf{mAP} & 
    \multirow{2}{*}{\textbf{U-Recall}} & \multicolumn{3}{c|}{\cellcolor{lightgray}\textbf{mAP}} &
    \multirow{2}{*}{\textbf{U-Recall}} & \multicolumn{3}{c|}{\cellcolor{lightgray}\textbf{mAP}} &
    \multicolumn{3}{c}{\cellcolor{lightgray}\textbf{mAP}}\\
     
    &  & \textbf{Curr.} & 
    & \textbf{Prev.} & \textbf{Curr.} &  \textbf{Both} &
    & \textbf{Prev.} & \textbf{Curr.} & \textbf{Both}
    & \textbf{Prev.} & \textbf{Curr.} & \textbf{Both} \\
    \midrule \midrule

    \multicolumn{14}{c}{\textbf{M-OWOD Benchmark Results}} \\ \midrule

    ORE~\cite{joseph2021ore}       & 4.9  & 56.0 & 2.9  & 52.7 & 26.0 & 39.4 & 3.9  & 38.2 & 12.7 & 29.7 & 29.6 & 12.4 & 25.3 \\
    OST~\cite{OST}       & -    & 56.2 & -    & 53.4 & 26.5 & 39.9 & -    & 38.0 & 12.8 & 29.6 & 30.1 & 13.3 & 25.9 \\
    OW-DETR~\cite{gupta2021ow}& 7.5  & 59.2 & 6.2  & 53.6 & 33.5 & 42.9 & 5.7  & 38.3 & 15.8 & 30.8 & 31.4 & 17.1 & 27.8 \\
    UC-OWOD~\cite{wu2022ucowod}& -    & 50.7 & -    & 33.1 & 30.5 & 31.8 & -    & 28.8 & 16.3 & 24.6 & 25.6 & 15.9 & 23.2 \\
    ALLOW~\cite{allow}   & 13.6 & 59.3 & 10.0 & 53.2 & 34.0 & 45.6 & 14.3 & 42.6 & 26.7 & 38.0 & 33.5 & 21.8 & 30.6 \\
    PROB~\cite{Zohar_2023_CVPR}     & 19.4 & 59.5 & 17.4 & 55.7 & 32.2 & 44.0 & 19.6 & 43.0 & 22.2 & 36.0 & 35.7 & 18.9 & 31.5 \\
    CAT~\cite{cat}       & 23.7 & 60.0 & 19.1 & 55.5 & 32.7 &44.1 & 24.4 & 42.8 & 18.7 & 34.8 & 34.4 & 16.6 & 29.9 \\
    RandBox~\cite{Randbox}& 10.6 & \textbf{61.8} & 6.3 & - & - & 45.3 & 7.8 & - & - & 39.4 & - & - & 35.4 \\

    OrthogonalDet~\cite{sun2024exploring} & 24.6 & 61.3 & 26.3 & 55.5 & 38.5 & 47.0 & 29.1 & 46.7 & 30.6 &  41.3 & 42.4 & 24.3 & 37.9 \\

    \rowcolor{LightGreen}
    \textbf{CROWD} (ours) & \textbf{57.9} & 61.7 & \textbf{53.6} & \textbf{56.7} & \textbf{38.9} & \textbf{47.8} & \textbf{69.6} & \textbf{48.0} & \textbf{31.4} & \textbf{42.5} & \textbf{42.9} & \textbf{25.4} & \textbf{38.5} \\
    
    \midrule
    \multicolumn{14}{c}{\textbf{S-OWOD}\textbf{ Benchmark Results}} \\ \midrule
    
    ORE~\cite{joseph2021ore}       & 1.5  & 61.4 & 3.9  & 56.5 & 26.1 & 40.6 & 3.6  & 38.7 & 23.7 & 33.7 & 33.6 & 26.3 & 31.8 \\
    OW-DETR~\cite{gupta2021ow}& 5.7  & 71.5 & 6.2  & 62.8 & 27.5 & 43.8 & 6.9  & 45.2 & 24.9 & 38.5 & 38.2 & 28.1 & 33.1 \\
    PROB~\cite{Zohar_2023_CVPR}     & 17.6 & 73.4 & 22.3 & 66.3 & 36.0 & 50.4 & 24.8 & 47.8 & 30.4 & 42.0 & 42.6 & 31.7 & 39.9 \\
    CAT~\cite{cat}       & 24.0 & \textbf{74.2} & 23.0 & \textbf{67.6} & 35.5 & 50.7 & 24.6 & 51.2 & 32.6 & 45.0 & 45.4 & 35.1 & 42.8 \\

    OrthogonalDet \cite{sun2024exploring} & 24.6 & 71.6 & 27.9 & 64.0 & 39.9 & 51.3 & 31.9 & 52.1 & \textbf{42.2} & \textbf{48.8} & 48.7 & 38.8 & 46.2 \\

    \rowcolor{LightGreen}
    \textbf{CROWD} (ours) & \textbf{50.5} & \textbf{73.5} & \textbf{41.7} & 64.9 & \textbf{41.2} & \textbf{53.1} & \textbf{49.6} & \textbf{54.7} & 42.1 & 48.4 & \textbf{49.8} & \textbf{43.0} & \textbf{46.4} \\
    \bottomrule
  \end{tabular}}%
  \label{tab:open_world_results}
  \vspace{-2ex}
\end{table*}

\begin{table*}[t]
\centering
\small
\caption{\textbf{Unknown Class Metrics on M-OWODB.} Comparison of U-Recall, WI, and A-OSE across tasks (excluding Task 4 where all classes are known $U^t = \phi$). Best results are in \textbf{bold}.}
\resizebox{\textwidth}{!}{%
\begin{tabular}{l|ccc|ccc|ccc}
\toprule
\textbf{Method} & 
\multicolumn{3}{c|}{\textbf{Task 1}} & 
\multicolumn{3}{c|}{\textbf{Task 2}} & 
\multicolumn{3}{c}{\textbf{Task 3}} \\
& \textbf{U-Recall ($\uparrow$)} & \textbf{WI ($\downarrow$)} & \textbf{A-OSE ($\downarrow$)} 
& \textbf{U-Recall ($\uparrow$)} & \textbf{WI ($\downarrow$)} & \textbf{A-OSE ($\downarrow$)} 
& \textbf{U-Recall ($\uparrow$)} & \textbf{WI ($\downarrow$)} & \textbf{A-OSE ($\downarrow$)} \\
\midrule
ORE~\cite{joseph2021ore}              & 4.9 & 0.0621 & 10459 & 2.9 & 0.0282 & 10445 & 3.9 & 0.0211 & 7990 \\
OST~\cite{OST}                        & -   & 0.0417 & 4889  & -   & 0.0213 & 2546  & -   & 0.0146 & 2120 \\
OW-DETR~\cite{gupta2021ow}            & 7.5 & 0.0571 & 10240 & 6.2 & 0.0278 & 8441 & 5.7 & 0.0156 & 6803 \\
PROB~\cite{Zohar_2023_CVPR}           & 19.4 & 0.0569 & 5195 & 17.4 & 0.0344 & 6452 & 19.6 & 0.0151 & 2641 \\
RandBox~\cite{Randbox}                & 10.6 & 0.0240 & 4498 & 6.3 & 0.0078 & 1880 & 7.8 & 0.0054 & 1452 \\
OrthogonalDet~\cite{sun2024exploring} & 24.6 & \textbf{0.0299} & 4148 & 26.3 & \textbf{0.0099} & 1791 & 29.1 & 0.0077 & 1345 \\
\midrule
\rowcolor{LightGreen}
\textbf{CROWD} (\textbf{Ours})        & \textbf{57.6} & 0.0380 & \textbf{3823} & \textbf{53.6} & 0.0101 & \textbf{1508} & \textbf{69.6} & \textbf{0.0066} & \textbf{1266}  \\
\bottomrule
\end{tabular}
} 
\label{tab:crowd_confusions}
\end{table*}

\noindent \textbf{Experimental Setup} : Following \citet{sun2024exploring} we adopt a Faster-RCNN~\cite{frcnn} based model with a pretrained ResNet-50~\cite{ResNet} backbone. 
Our model is trained incrementally on 4 tasks as described above with a batch size of 12, an AdamW optimizer, a base learning rate to $2.5 \times 10^{-5}$ and weight decay of $1 \times 10^{-4}$.
CROWD-D utilizes the RoI features ($|\mathtt{R}| = 500$) to mine $\mathtt{k}=10$ unknown instances (determined through ablation study in \cref{app:ablation_budget}) per image. The CROWD-L loss is applied across tasks as an additional head and operates on RoI features projected to a 256-dimensional feature space. We train our model on 4 NVIDIA V100 GPUs, provide additional experimental details in \cref{app:additional_expts} and release our code at \url{https://github.com/amajee11us/CROWD.git}.\looseness-1

\noindent \textbf{Metrics} : We use mean average precision (mAP) to evaluate known classes, partitioned into previously seen and newly introduced categories. 
For unknown object class, we follow OWOD conventions~\cite{joseph2021ore,gupta2021ow} and report unknown object recall (U-Recall), as mAP is inapplicable due to incomplete annotations. 
To measure confusion between known and unknown classes, we report Wilderness Impact (WI)~\cite{wi_ref} and Absolute Open-Set Error (A-OSE)~\cite{ose_ref}.\looseness-1

\subsection{Results on Benchmark OWOD and IOD tasks}
\label{sec:benchmark_results}
\noindent \textbf{OWOD}: We compare the performance of CROWD against several existing baselines on M-OWOD and S-OWOD benchmarks as shown in \cref{tab:open_world_results}. Note, that we follow \citet{sun2024exploring} and report our results on the same seed and compute settings for fair comparisons.
\begin{wraptable}[25]{r}{0.4\textwidth} 
  \vspace{-10pt}
  \small
  \centering
  \caption{\small \textbf{Results of CROWD on PASCAL VOC for three IOD tasks} shown in terms of Prev., Curr., and overall mAP.}
  \label{tab:iod_summary}
  \resizebox{0.4\textwidth}{!}{\begin{tabular}{@{}lccc@{}}
    \toprule
    \multicolumn{4}{c}{\textbf{10 + 10 setting}} \\ 
    \midrule \midrule
    & \textbf{Prev.} & \textbf{Curr.} & \textbf{mAP} \\ 
    \midrule
    ILOD \cite{ILOD}                      & 63.2 & 63.2 & 63.2 \\
    Faster ILOD \cite{F_ILOD}             & 69.8 & 54.5 & 62.1 \\
    PROB~\cite{Zohar_2023_CVPR}           & 66.0 & 67.2 & 66.5 \\
    OrthogonalDet~\cite{sun2024exploring} & 69.4 & 71.8 & 67.0 \\
    \midrule
    \textbf{CROWD (ours)}                 & \textbf{73.5} & \textbf{75.1} & \textbf{72.0} \\
    \midrule
    \multicolumn{4}{c}{\textbf{15 + 5 setting}} \\ 
    \midrule
    \midrule
    ILOD \cite{ILOD}                      & 68.3 & 58.4 & 65.8 \\
    Faster ILOD \cite{F_ILOD}             & 71.6 & 56.9 & 67.9 \\
    PROB~\cite{Zohar_2023_CVPR}           & 73.2 & 60.8 & 70.1 \\
    OrthogonalDet~\cite{sun2024exploring} & 74.5 & 66.9 & 72.6 \\
    \midrule
    \textbf{CROWD (ours)}                 & \textbf{76.2} & \textbf{68.9} & \textbf{74.4} \\
    \midrule
    \multicolumn{4}{c}{\textbf{19 + 1 setting}} \\ 
    \midrule
    \midrule
    ILOD \cite{ILOD}                      & 68.5 & 62.7 & 68.2 \\
    Faster ILOD \cite{F_ILOD}             & 68.9 & 61.1 & 68.5 \\
    PROB~\cite{Zohar_2023_CVPR}           & 73.9 & 48.5 & 72.6 \\
    OrthogonalDet~\cite{sun2024exploring} & 73.5 & 74.5 & 73.6 \\
    \midrule
    \textbf{CROWD (ours)}                 & \textbf{74.2} & \textbf{75.3} & \textbf{74.2} \\
    \bottomrule
  \end{tabular}}
  \vspace{-6pt}
\end{wraptable}

\begin{figure*}
        \centering
        \begin{tabular}{lccc}
            \rotatebox{90}{\small OrthogonalDet~\cite{sun2024exploring}} &
                \includegraphics[width=0.25\textwidth]{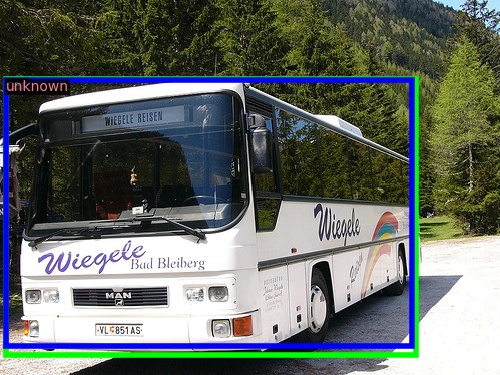} & 
                \includegraphics[width=0.25\textwidth]{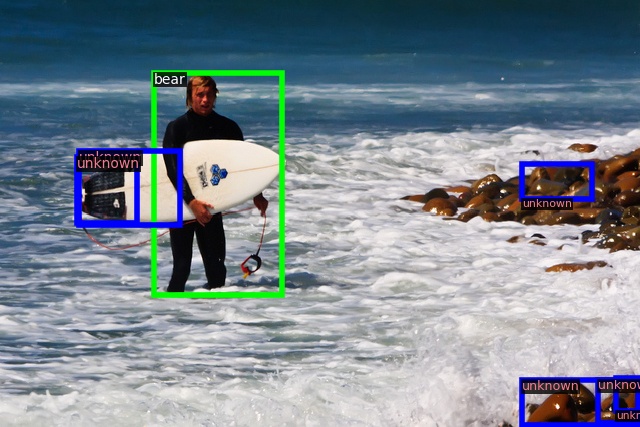} & 
                \includegraphics[width=0.25\textwidth]{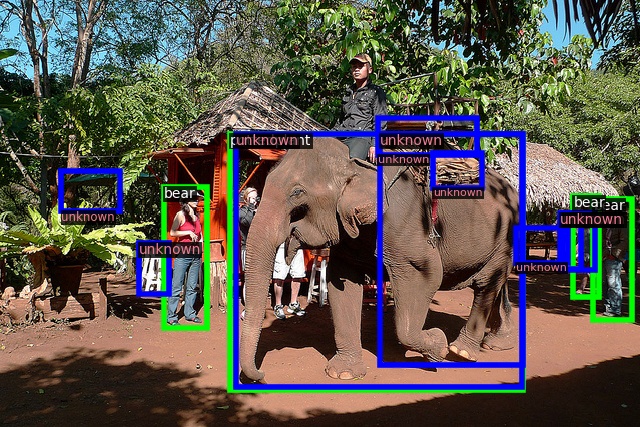} \\
            \rotatebox{90}{\small CROWD (ours) } & 
                \includegraphics[width=0.25\textwidth]{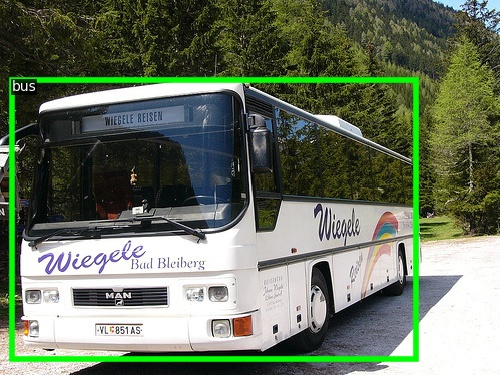} & 
                \includegraphics[width=0.25\textwidth]{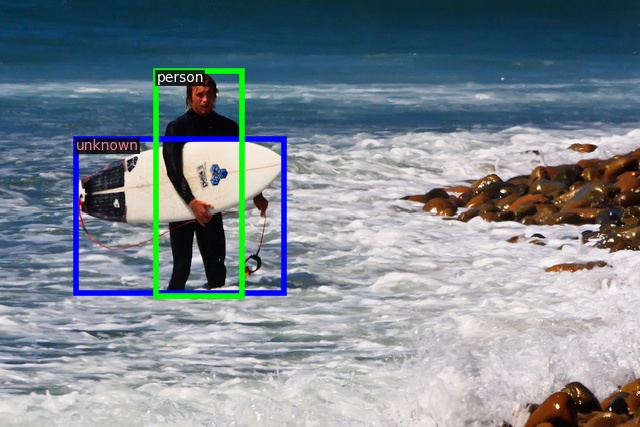} & 
                \includegraphics[width=0.25\textwidth]{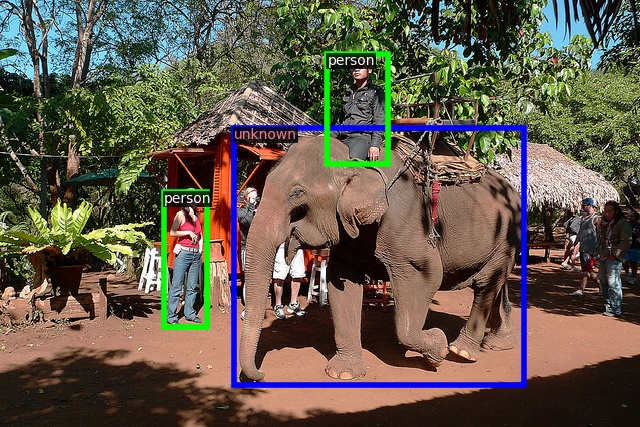} \\
            & (a)  & (b)  & (c) \\
        \end{tabular}
        \caption{\textbf{Qualitative results from CROWD} contrasted against OrthogonalDet~\cite{sun2024exploring} showing that our approach mitigates (a) confusion (b) generalizes to unknowns and (c) reduces forgetting.}
        \label{fig:qual}
        \vspace{-2ex}
\end{figure*}

CROWD surpasses the latest baseline OrthogonalDet~\cite{sun2024exploring} by up to 2.8\% and 2.1\% on M-OWOD and S-OWOD benchmarks while achieving up to 2.4$\times$ gains in U-recall. 
For approaches like PROB~\cite{Zohar_2023_CVPR}, CAT~\cite{cat} which adopt selection strategies to mine unknowns our combinatorial approach achieves up to 8.4\% (on M-OWOD) improvements.
This can be attributed to the contributions of CROWD-D which mines representative unknown examples effectively increasing the coverage on such objects.
Also, we observe $\sim3$\% increase in mAP for previously known classes indicating a reduction in forgetting. The competitive results on the currently known classes (Curr. in \cref{tab:open_world_results}) indicates that $h^t(.;\theta)$ enforces a stronger decision boundary between $K^t$ and $U^t$ through $L_{\text{CROWD}}^{cross}$ while retaining performance on $K^t$ through $L_{\text{CROWD}}^{self}$.
Additionally, in \cref{tab:crowd_confusions} we show that CROWD achieves lesser confusion over existing baselines while boosting U-Recall establishing the importance of modeling OWOD as a combinatorial data-discovery problem. This is further highlighted qualitatively in \cref{fig:qual}.\looseness-1

\noindent \textbf{IOD}: Our novel loss formulation described in \cref{sec:crowd_l} (point 4) is applied to the finetuning stage of IOD across three popular task splits from the PASCAL-VOC~\cite{Everingham15} dataset. Note, that for IOD we do not apply CROWD-D due to absence of unknown examples. 
Our results summarized in \cref{tab:iod_summary} and detailed in \cref{tab:iod_benchmark} (Appendix) shows up to 5.9\% boost in overall mAP showing \textit{better generalization to IOD} tasks while \textit{minimizing the impact of forgetting via stronger retention of previously known classes}, a very common pitfall in IOD.\looseness-1

\begin{table*}[t]
  \centering
  \caption{\textbf{Ablation Experiments on the M-OWOD benchmark.} We report the U-Recall and mAP (all known classes) by varying the choice of selection strategies in CROWD-D and learning objectives in CROWD-L. We show that a joint (data discovery + combinatorial loss) strategy provides the best overall performance (denoted as CROWD (joint)).}
  \resizebox{\textwidth}{!}{%
  \begin{tabular}{l|ccc|cc|cc|cc|c}
    \toprule
    \textbf{Method} & Baseline & CROWD & CROWD &
    \multicolumn{2}{c|}{\textbf{Task 1}} & 
    \multicolumn{2}{c|}{\textbf{Task 2}} & 
    \multicolumn{2}{c|}{\textbf{Task 3}} & 
    \textbf{Task 4} \\ 
    
    & & -D & -L & 
    \textbf{U-Recall} & \textbf{mAP} & 
    \textbf{U-Recall} & \textbf{mAP} & 
    \textbf{U-Recall} & \textbf{mAP} & 
    \textbf{mAP}  \\
    \midrule \midrule

    OrthogonalDet \cite{sun2024exploring} & \checkmark &  &  & 24.6\textsubscript{$\pm$0.04} & 61.3\textsubscript{$\pm$0.11} & 26.3\textsubscript{$\pm$0.01} & 47.0\textsubscript{$\pm$0.06} & 29.1\textsubscript{$\pm$0.01} & 41.3\textsubscript{$\pm$0.10} & 37.9\textsubscript{$\pm$0.09} \\ \midrule
    
    CROWD-D (w/ FLCG)  & \checkmark & \checkmark &    & 50.7\textsubscript{$\pm$0.23} & 60.3\textsubscript{$\pm$0.07} & 52.2\textsubscript{$\pm$0.33} & 45.7\textsubscript{$\pm$0.04} & 60.1\textsubscript{$\pm$0.18} & 40.6\textsubscript{$\pm$0.03} & 38.3\textsubscript{$\pm$0.11} \\
    \rowcolor{LightCyan}
    CROWD-D (w/ GCCG) & \checkmark & \checkmark &     & 57.0\textsubscript{$\pm$0.17} & 61.2\textsubscript{$\pm$0.05} & 54.1\textsubscript{$\pm$0.72} & 45.2\textsubscript{$\pm$0.02} & 69.6\textsubscript{$\pm$0.11} & 40.8\textsubscript{$\pm$0.01} &  38.1\textsubscript{$\pm$0.09} \\
    CROWD-D (w/ LogDetCG) & \checkmark & \checkmark & & 56.4\textsubscript{$\pm$0.46} & 61.2\textsubscript{$\pm$0.10} & 54.1\textsubscript{$\pm$0.65} & 44.1\textsubscript{$\pm$0.07} & 69.1\textsubscript{$\pm$0.26} & 39.7\textsubscript{$\pm$0.10} &  37.6\textsubscript{$\pm$0.08}   \\ \midrule

    \rowcolor{LightCyan}
    CROWD-L (w/ FLCG)  & \checkmark & & \checkmark    & 25.0\textsubscript{$\pm$0.01} & 61.7\textsubscript{$\pm$0.02} & 26.8\textsubscript{$\pm$0.03} & 47.7\textsubscript{$\pm$0.16} & 28.8\textsubscript{$\pm$0.30} & 42.4\textsubscript{$\pm$0.11} & 38.5\textsubscript{$\pm$0.06}  \\
    CROWD-L (w/ GCCG) & \checkmark & & \checkmark     & 24.3\textsubscript{$\pm$0.03} & 61.3\textsubscript{$\pm$0.12} & 27.1\textsubscript{$\pm$0.10} & 47.4\textsubscript{$\pm$0.26} & 31.0\textsubscript{$\pm$0.44} & 40.2\textsubscript{$\pm$0.11} & 38.2\textsubscript{$\pm$0.10}   \\
    CROWD-L (w/ LogDetCG) & \checkmark & & \checkmark & 22.7\textsubscript{$\pm$0.01} & 59.5\textsubscript{$\pm$0.09} & 27.0\textsubscript{$\pm$0.14} & 44.6\textsubscript{$\pm$0.22} & 27.2\textsubscript{$\pm$0.21} & 38.3\textsubscript{$\pm$0.14} & 36.0\textsubscript{$\pm$0.27}   \\ \midrule

    \rowcolor{LightGreen}
    CROWD (joint) & \checkmark & \checkmark & \checkmark & $57.9_{\pm0.33}$ & $61.7_{\pm0.02}$ & $53.6_{\pm0.41}$ & $47.8_{\pm0.02}$ & $69.6_{\pm0.26}$ & $31.4_{\pm0.03}$ & $38.5_{\pm0.07}$ \\ 
    \bottomrule
  \end{tabular}}
  \label{tab:owod_ablations}
  \vspace{-2ex}
\end{table*}

\subsection{Ablations}
\label{sec:all_ablations}
We conduct ablations on the M-OWOD benchmark to analyze the contributions of individual components of CROWD. 
On top of the baseline method OrthogonalDet~\cite{sun2024exploring} we first introduce instances of CROWD-D to assess the impact of data-discovery under a fixed budget $\mathtt{k} = 10$.
Next, we decouple CROWD-D and introduce our novel learning objectives in CROWD-L to assess their impact on forgetting and confusion as discussed in \cref{sec:crowd_l}. For each of the above steps we ablate among instances of $f$ - Graph-Cut (GC), Log-Determinant (LogDet) and Facility-Location (FL). Finally, we combine the best performing instances from CROWD-D and CROWD-L into a \emph{joint} formulation (referred to as CROWD (joint)) as shown in \cref{tab:owod_ablations} which \textit{achieves the best overall performance, balancing the tradeoff between boosting currently known class performance and retaining performance on previously learnt ones}.\looseness-1

\noindent \textbf{Impact of Data-Discovery in CROWD-D} : As shown in \cref{tab:owod_ablations}, irrespective of the choice of $f$, CROWD-D boosts the U-Recall over the baseline by introducing additional information in the form of pseudo labeled unknowns.
We observe that CROWD-D (w/GCCG) ($f$ here is Graph-Cut) provides the best gains in U-Recall up to 2$\times$ over the latest baseline OrthogonalDet. 
This follows the observation in \citet{prism} which shows that greedy maximization of GCCG models relevance (examples which are dissimilar to both $K^t$ and $U^t$) while others model diversity (CROWD-D w/LogDet) and representation (CROWD-D w/Facility-Location). Thus, we adopt GCCG based selection strategy in \cref{alg:crowd_discover} for our experiments in \cref{tab:open_world_results}.\looseness-1

\noindent \textbf{Impact of $\mathtt{k}$} : As stated in \cref{sec:crowd_d}, $\mathtt{k}$ controls the number of potential unknown RoIs identified by CROWD-D per image. We ablate among several plausible values of $\mathtt{k} \in [0, 100]$ and summarize the results in \cref{tab:ablation_k} of the Appendix. 
Increasing the number of identified unknowns from 0 (OrthogonalDet) to 10 shows an increase in performance of the underlying model (U-Recall) while the performance does not increase beyond 20.
The increase in U-Recall can be attributed to inclusion of informative RoIs in the training loop.
In fact, the mAP on known classes slightly drops below existing baselines for $\mathtt{k} = 100$ due to inclusion of spurious background RoIs in the training pipeline.\looseness-1

\newcommand{\centered}[1]{\begin{tabular}{l} #1 \end{tabular}}

\noindent \textbf{Impact of Combinatorial Objectives in CROWD-L} : Similar to CROWD-D we ablate on variations of $f$ to contrast between formulations summarized in \cref{tab:crowd_instances}. As shown in \cref{tab:owod_ablations} our learning formulation, particularly CROWD-FL (based on Facility-Location) demonstrates better retention of previously known class performance while achieving competitive results on latest baseline OrthogonalDet.
This follows the observation in \citet{score} which demonstrates that FL based objectives model representation, retaining the most discriminative features through $L_{\text{CROWD}}^{self}$ while enforcing sufficient inter-cluster boundary between known and unknown RoI features ($L_{\text{CROWD}}^{cross}$). 
This also re-establishes the properties described in \cref{fig:crowd_loss_characterization} wherein CROWD-FL shows larger sensitivity to inter-cluster separation as compared to CROWD-GC, CROWD-LogDet and $L_{decorr}$ introduced in OrthogonalDet.\looseness-1

\begin{table*}[t]
  \centering
  \caption{\textbf{Ablation Experiments on the exclusion criterion $\tau_e$ and background budget $\tau_b$ in CROWD-D.} The submodular function is kept constant as Graph-Cut in CROWD-D and a CROWD-FL based learning objective is chosen in CROWD-L under constant machine seed.}
  \resizebox{0.8\textwidth}{!}{%
  \begin{tabular}{l|c|cc|cc|cc|c}
    \toprule
    \textbf{Method} & Value &
    \multicolumn{2}{c|}{\textbf{Task 1}} & 
    \multicolumn{2}{c|}{\textbf{Task 2}} & 
    \multicolumn{2}{c|}{\textbf{Task 3}} & 
    \textbf{Task 4} \\ 
    
    &  & 
    \textbf{U-Recall} & \textbf{mAP} & 
    \textbf{U-Recall} & \textbf{mAP} & 
    \textbf{U-Recall} & \textbf{mAP} & 
    \textbf{mAP}  \\
    \midrule \midrule
    
    \multirow{3}{*}{\textbf{CROWD ($\tau_e$)}}   & 0.05 & 
                                                   57.5 & 61.7 & 
                                                   52.7 & 47.1 & 
                                                   68.2 & 41.5  & 
                                                   38.0 \\
                    & \textbf{0.2}  & 
                      \cellcolor{LightGreen}$57.9$ & \cellcolor{LightGreen}$61.7$ & \cellcolor{LightGreen}$53.6$ & \cellcolor{LightGreen}$47.8$ & \cellcolor{LightGreen}$69.6$ & \cellcolor{LightGreen}$42.4$ & \cellcolor{LightGreen}$38.5$ \\
          $\tau_b = 30\%$ & 0.5  & 
                    53.3 & 59.5 & 
                    49.4 & 45.9 & 
                    65.0 & 40.1 & 
                    37.8  \\ 
    \midrule
     \multirow{3}{*}{\textbf{CROWD ($\tau_b$)}}   & 10\% & 
                                                    57.9 & 61.7 & 
                                                    53.1 & 46.8 & 
                                                    69.2 & 41.3  & 
                                                    37.4 \\
                  & \textbf{30}\%  & 
                    \cellcolor{LightGreen}$57.9$ & \cellcolor{LightGreen}$61.7$ & \cellcolor{LightGreen}$53.6$ & \cellcolor{LightGreen}$47.8$ & \cellcolor{LightGreen}$69.6$ & \cellcolor{LightGreen}$42.4$ & \cellcolor{LightGreen}$38.5$ \\
        $\tau_e = 0.2$ & 50\%  & 
                    55.4 & 60.0 & 
                    50.0 & 42.1 & 
                    63.7 & 38.9 & 
                    34.5  \\ 
     \bottomrule
  \end{tabular}}
  \label{tab:ablation_tau_e_tau_b}
\end{table*}

\noindent \textbf{Ablation on Exclusion Criterion $\tau_e$ and $\tau_b$ in CROWD-D} - At first, $\tau_e$ is an exclusion threshold which reduces the search space of CROWD-D by eliminating RoIs which have a low confidence threshold. As shown in \cref{tab:ablation_tau_e_tau_b}, increasing $\tau_e$ from 0 to 1 increases performance until $\tau_e = 0.2$ and then reduces. A lower value of $\tau_e$ allows for a large search space but includes a lot of noisy background objects leading to reduced selection performance. On the other hand a large value of $\tau_e$ can potentially earmark unknown foregrounds as unknowns resulting in reduced performance.

Keeping $\tau_e$ fixed at 0.2 we ablate $\tau_b$ which controls the selection budget for backgrounds (higher the value more are the number of background RoIs identified). Increasing $\tau_b$ (percentage here) increases the fraction of RoIs treated as backgrounds. This widens the search space for the combinatorial function causing a small drop in performance due to confusions between true backgrounds and foreground unknowns. On the other hand, very large values of $\tau_b$ shrink the search space oftentimes considering unknown foregrounds as background objects showing a steep drop in performance.

\begin{table*}[t]
  \centering
  \caption{\textbf{Ablation Experiments on the variation in $\eta$ in CROWD-L.} Given the Graph-Cut based selection strategy in CROWD-D we vary $\eta$ between [0.5, 1.0, 1.5] and adopt the best performing value for our pipeline in CROWD-L. The selection budget $\mathtt{k}$ in CROWD-D was set to 10 for all experiments and a fixed seed value.}
   \resizebox{\textwidth}{!}{%
  \begin{tabular}{l|c|cc|cccc|cccc|ccc}
    \toprule
    \multirow{3}{*}{\textbf{Method}} & 
    \multirow{3}{*}{$\eta$} &
    \multicolumn{2}{c|}{\textbf{Task 1}} & 
    \multicolumn{4}{c|}{\textbf{Task 2}} & 
    \multicolumn{4}{c|}{\textbf{Task 3}} & 
    \multicolumn{3}{c}{\textbf{Task 4}} \\ \cline{3 - 15} 

    & & \multirow{2}{*}{\textbf{U-Recall}} & \cellcolor{lightgray}\textbf{mAP} & 
    \multirow{2}{*}{\textbf{U-Recall}} & \multicolumn{3}{c|}{\cellcolor{lightgray}\textbf{mAP}} &
    \multirow{2}{*}{\textbf{U-Recall}} & \multicolumn{3}{c|}{\cellcolor{lightgray}\textbf{mAP}} &
    \multicolumn{3}{c}{\cellcolor{lightgray}\textbf{mAP}}\\
     
    & &  & \textbf{Curr.} & 
    & \textbf{Prev.} & \textbf{Curr.} &  \textbf{Both} &
    & \textbf{Prev.} & \textbf{Curr.} & \textbf{Both}
    & \textbf{Prev.} & \textbf{Curr.} & \textbf{Both} \\
    \midrule \midrule

    OrthogonalDet \cite{sun2024exploring} & - & 24.6 & 61.3 & 
                                                26.3 & 55.5 & 38.5 & 47.0 & 
                                                29.1 & 46.7 & 30.6 & 41.3 & 
                                                       42.4 & 24.3 & 37.9 \\ \midrule
    
    \multirow{3}{*}{\textbf{CROWD (ours)}}   & 0.5 & 57.8 & 58.8 & 
                                                     53.4 & 57.1 & 32.8 & 44.9 & 
                                                     65.3 & 50.2 & 25.9 & 42.1 & 
                                                            44.0 & 21.1 & 38.3 \\
     & \textbf{1.0} & \cellcolor{LightGreen}\textbf{57.9} & \cellcolor{LightGreen}\textbf{61.7} & 
                       \cellcolor{LightGreen}\textbf{53.6} & \cellcolor{LightGreen}\textbf{56.7} & \cellcolor{LightGreen}\textbf{38.9} & \cellcolor{LightGreen}\textbf{47.8} & 
                       \cellcolor{LightGreen}\textbf{69.6} & \cellcolor{LightGreen}\textbf{48.0} & \cellcolor{LightGreen}\textbf{31.4} & \cellcolor{LightGreen}\textbf{42.5} & 
                       \cellcolor{LightGreen}\textbf{42.9} & \cellcolor{LightGreen}\textbf{25.4} & \cellcolor{LightGreen}\textbf{38.5} \\
     & 1.5  & 57.9 & 61.7 & 
              53.6 & 55.6 & 39.1 & 47.4 & 
              69.5 & 44.0 & 34.6 & 40.9 & 
                     44.0 & 21.1 & 38.3 \\ 
    \bottomrule
  \end{tabular}}
  \label{tab:ablation_eta}
\end{table*}

\noindent \textbf{Ablation on Trade-off between $L_{\text{CROWD}}^{self}$ and $L_{\text{CROWD}}^{cross}$ in CROWD-L} - The hyper-parameter $\eta$ controls the trade-off between known-unknown class separation and known class cluster compactness discussed in \cref{tab:ablation_eta}. A lower value of $\eta$ does not enforce separation between currently known and unknown exemplars but enforces intra-class compactness. 
This results in better retention of previously known objects but a drop in currently known objects due to increased confusion with unknown exemplars. On the other hand for a large value of $\eta$ the model enforces large separation between currently known and unknown objects boosting performance on the currently knowns but suffers from catastrophic forgetting of the previously known classes.

\vspace{-2ex}
\section{Conclusion, Limitations and Future Work}
\vspace{-1ex}
\label{sec:conclusion}
We introduced CROWD, a novel combinatorial framework in OWOD, which reformulates OWOD as interleaved set-based discovery (CROWD-D) and representation learning (CROWD-L) tasks. 
Leveraging Submodular Conditional Gain (SCG) functions, CROWD-D strategically selects representative unknown instances distinctly dissimilar from known objects while CROWD-L consumes mined unknowns to preserve discriminative coherence over known classes. 
Our evaluations confirm that CROWD effectively addresses known vs. unknown class confusion and forgetting, achieving significant improvements in unknown recall and known-class accuracy on standard OWOD and IOD benchmarks. 
Despite the significant improvements in U-Recall CROWD-D continues to inject a small set of spurious exemplars into the selected pool which we aim to address in future works by exploring alternative combinatorial formulations beyond SCG, and introducing stricter constraints in CROWD-D.\looseness-1

\section*{Acknowledgements}
We gratefully thank anonymous reviewers for their valuable comments. We would also like to extend our gratitude to our fellow researchers from the CARAML lab at UT Dallas for their suggestions. This work is supported by the National Science Foundation under Grant Numbers IIS-2106937, a gift from Google Research, an Amazon Research Award, and the Adobe Data Science Research award. Any opinions, findings, and conclusions or recommendations expressed in this material are those of the authors and do not necessarily reflect the views of the National Science Foundation, Google or Adobe.

{
    \small
    \bibliographystyle{plainnat}
    \bibliography{references}

\begin{thebibliography}{75}
\providecommand{\natexlab}[1]{#1}
\providecommand{\url}[1]{\texttt{#1}}
\expandafter\ifx\csname urlstyle\endcsname\relax
  \providecommand{\doi}[1]{doi: #1}\else
  \providecommand{\doi}{doi: \begingroup \urlstyle{rm}\Url}\fi

\bibitem[Aljundi et~al.(2019)Aljundi, Lin, Goujaud, and Bengio]{Gradient_based_sample_selection}
Rahaf Aljundi, Min Lin, Baptiste Goujaud, and Yoshua Bengio.
\newblock Gradient based sample selection for online continual learning.
\newblock In \emph{Advances in Neural Information Processing Systems}, volume~32, 2019.

\bibitem[Beck et~al.(2021)Beck, Sivasubramanian, Dani, Ramakrishnan, and Iyer]{Beck2021EffectiveEO}
Nathan Beck, Durga Sivasubramanian, Apurva Dani, Ganesh Ramakrishnan, and Rishabh~K. Iyer.
\newblock Effective evaluation of deep active learning on image classification tasks.
\newblock \emph{ArXiv}, abs/2106.15324, 2021.

\bibitem[Carion et~al.(2020)Carion, Massa, Synnaeve, Usunier, Kirillov, and Zagoruyko]{DETR}
Nicolas Carion, Francisco Massa, Gabriel Synnaeve, Nicolas Usunier, Alexander Kirillov, and Sergey Zagoruyko.
\newblock End-to-end object detection with transformers, 2020.

\bibitem[Coleman et~al.(2020)Coleman, Yeh, Mussmann, Mirzasoleiman, Bailis, Liang, Leskovec, and Zaharia]{SelectionviaProxy}
Cody Coleman, Christopher Yeh, Stephen Mussmann, Baharan Mirzasoleiman, Peter Bailis, Percy Liang, Jure Leskovec, and Matei Zaharia.
\newblock Selection via proxy: Efficient data selection for deep learning.
\newblock In \emph{International Conference on Learning Representations}, 2020.

\bibitem[Dai et~al.(2021)Dai, Cai, Lin, and Chen]{UP-DETR}
Zhigang Dai, Bolun Cai, Yugeng Lin, and Junying Chen.
\newblock Up-detr: Unsupervised pre-training for object detection with transformers.
\newblock In \emph{2021 IEEE/CVF Conference on Computer Vision and Pattern Recognition (CVPR)}, pages 1601--1610, 2021.

\bibitem[Dhamija et~al.(2020)Dhamija, Gunther, Ventura, and Boult]{wi_ref}
Akshay Dhamija, Manuel Gunther, Jonathan Ventura, and Terrance Boult.
\newblock The overlooked elephant of object detection: Open set.
\newblock In \emph{Proceedings of the IEEE/CVF Winter Conference on Applications of Computer Vision (WACV)}, March 2020.

\bibitem[Duan et~al.(2019)Duan, Bai, Xie, Qi, Huang, and Tian]{centernet}
Kaiwen Duan, Song Bai, Lingxi Xie, Honggang Qi, Qingming Huang, and Qi~Tian.
\newblock Centernet: Keypoint triplets for object detection.
\newblock In \emph{2019 IEEE/CVF International Conference on Computer Vision (ICCV)}, pages 6568--6577, 2019.

\bibitem[Ducoffe and Precioso(2018)]{ducoffe2018adversarial}
Melanie Ducoffe and Frederic Precioso.
\newblock Adversarial active learning for deep networks: a margin based approach.
\newblock \emph{arXiv preprint arXiv:1802.09841}, 2018.

\bibitem[Everingham et~al.(2015)Everingham, Eslami, Van~Gool, Williams, Winn, and Zisserman]{Everingham15}
M.~Everingham, S.~M.~A. Eslami, L.~Van~Gool, C.~K.~I. Williams, J.~Winn, and A.~Zisserman.
\newblock The pascal visual object classes challenge: A retrospective.
\newblock \emph{International Journal of Computer Vision}, 111\penalty0 (1):\penalty0 98--136, January 2015.

\bibitem[Fujishige(2005)]{fujishige}
Satoru Fujishige.
\newblock \emph{Submodular Functions and Optimization}, volume~58.
\newblock Elsevier, 2005.

\bibitem[Fukuda et~al.(2024)Fukuda, Kera, and Kawamoto]{fukuda2024adapter}
Takuma Fukuda, Hiroshi Kera, and Kazuhiko Kawamoto.
\newblock Adapter merging with centroid prototype mapping for scalable class-incremental learning, 2024.

\bibitem[Girshick(2015)]{FastRCNN}
Ross Girshick.
\newblock Fast r-cnn.
\newblock In \emph{2015 IEEE International Conference on Computer Vision (ICCV)}, pages 1440--1448, 2015.

\bibitem[Girshick et~al.(2014)Girshick, Donahue, Darrell, and Malik]{rcnn}
Ross Girshick, Jeff Donahue, Trevor Darrell, and Jitendra Malik.
\newblock Rich feature hierarchies for accurate object detection and semantic segmentation.
\newblock CVPR '14, page 580–587, 2014.

\bibitem[Gupta et~al.(2022)Gupta, Narayan, Joseph, Khan, Khan, and Shah]{gupta2021ow}
Akshita Gupta, Sanath Narayan, KJ~Joseph, Salman Khan, Fahad~Shahbaz Khan, and Mubarak Shah.
\newblock Ow-detr: Open-world detection transformer.
\newblock In \emph{CVPR}, 2022.

\bibitem[He et~al.(2016)He, Zhang, Ren, and Sun]{ResNet}
Kaiming He, Xiangyu Zhang, Shaoqing Ren, and Jian Sun.
\newblock Deep residual learning for image recognition.
\newblock In \emph{2016 IEEE Conference on Computer Vision and Pattern Recognition (CVPR)}, pages 770--778, 2016.

\bibitem[Iyer et~al.(2022)Iyer, Khargonkar, Bilmes, and Asnani]{alt_gen_submod}
Rishabh Iyer, Ninad Khargonkar, Jeff Bilmes, and Himanshu Asnani.
\newblock Generalized submodular information measures: Theoretical properties, examples, optimization algorithms, and applications.
\newblock \emph{IEEE Transactions on Information Theory}, 68\penalty0 (2):\penalty0 752--781, 2022.

\bibitem[Jain et~al.(2023)Jain, Nandy, Aggarwal, Tendulkar, Iyer, and De]{jain2023efficient}
Eeshaan Jain, Tushar Nandy, Gaurav Aggarwal, Ashish~V. Tendulkar, Rishabh~K Iyer, and Abir De.
\newblock Efficient data subset selection to generalize training across models: Transductive and inductive networks.
\newblock In \emph{Thirty-seventh Conference on Neural Information Processing Systems ({NeurIPS})}, 2023.

\bibitem[Jegelka and Bilmes(2011)]{submod_cooperation}
Stefanie Jegelka and Jeff Bilmes.
\newblock Submodularity beyond submodular energies: Coupling edges in graph cuts.
\newblock In \emph{CVPR 2011}, 2011.

\bibitem[Joseph et~al.(2021)Joseph, Khan, Khan, and Balasubramanian]{joseph2021ore}
K~J Joseph, Salman Khan, Fahad~Shahbaz Khan, and Vineeth~N Balasubramanian.
\newblock Towards open world object detection.
\newblock In \emph{2021 IEEE/CVF Conference on Computer Vision and Pattern Recognition (CVPR)}, page 5826–5836, jun 2021.

\bibitem[Joseph et~al.(2022)Joseph, Rajasegaran, Khan, Khan, and Balasubramanian]{Meta_ILOD}
K.~J. Joseph, Jathushan Rajasegaran, Salman Khan, Fahad~Shahbaz Khan, and Vineeth~N. Balasubramanian.
\newblock Incremental object detection via meta-learning.
\newblock \emph{IEEE Transactions on Pattern Analysis and Machine Intelligence}, 44\penalty0 (12):\penalty0 9209--9216, 2022.

\bibitem[Karanam et~al.(2022)Karanam, Killamsetty, Kokel, and Iyer]{orient}
Athresh Karanam, Krishnateja Killamsetty, Harsha Kokel, and Rishabh Iyer.
\newblock Orient: Submodular mutual information measures for data subset selection under distribution shift.
\newblock In \emph{Advances in Neural Information Processing Systems}, volume~35, 2022.

\bibitem[Kaushal et~al.(2019{\natexlab{a}})Kaushal, Iyer, Doctor, Sahoo, Dubal, Kothawade, Mahadev, Dargan, and Ramakrishnan]{vid_sum_2019}
V.~Kaushal, R.~Iyer, K.~Doctor, A.~Sahoo, P.~Dubal, S.~Kothawade, R.~Mahadev, K.~Dargan, and G.~Ramakrishnan.
\newblock Demystifying multi-faceted video summarization: Tradeoff between diversity, representation, coverage and importance.
\newblock In \emph{2019 IEEE Winter Conference on Applications of Computer Vision (WACV)}, pages 452--461, 2019{\natexlab{a}}.

\bibitem[Kaushal et~al.(2019{\natexlab{b}})Kaushal, Iyer, Kothawade, Mahadev, Doctor, and Ramakrishnan]{Kaushal_2019}
Vishal Kaushal, Rishabh Iyer, Suraj Kothawade, Rohan Mahadev, Khoshrav Doctor, and Ganesh Ramakrishnan.
\newblock Learning from less data: A unified data subset selection and active learning framework for computer vision.
\newblock In \emph{2019 IEEE Winter Conference on Applications of Computer Vision (WACV)}, 2019{\natexlab{b}}.

\bibitem[Kaushal et~al.(2019{\natexlab{c}})Kaushal, Subramanian, Kothawade, Iyer, and Ramakrishnan]{kaushal2019framework}
Vishal Kaushal, Sandeep Subramanian, Suraj Kothawade, Rishabh Iyer, and Ganesh Ramakrishnan.
\newblock A framework towards domain specific video summarization.
\newblock In \emph{2019 IEEE winter conference on applications of computer vision (WACV)}, pages 666--675. IEEE, 2019{\natexlab{c}}.

\bibitem[Killamsetty et~al.(2024)Killamsetty, Abhishek, Aakriti, Ramakrishnan, Evfimievski, Popa, and Iyer]{killamsetty_automata}
Krishnateja Killamsetty, Guttu~Sai Abhishek, Aakriti, Ganesh Ramakrishnan, Alexandre~V. Evfimievski, Lucian Popa, and Rishabh Iyer.
\newblock {AUTOMATA}: gradient based data subset selection for compute-efficient hyper-parameter tuning.
\newblock In \emph{Proceedings of the 36th International Conference on Neural Information Processing Systems}, 2024.

\bibitem[Kim et~al.(2021)Kim, Lin, Angelova, Kweon, and Kuo]{kim2021openproposals}
Dahun Kim, Tsung-Yi Lin, Anelia Angelova, In~So Kweon, and Weicheng Kuo.
\newblock Learning open-world object proposals without learning to classify.
\newblock \emph{arXiv preprint arXiv:2104.05704}, 2021.

\bibitem[Knoblauch et~al.(2020)Knoblauch, Husain, and Diethe]{OptimalContinualLearning}
Jeremias Knoblauch, Hisham Husain, and Tom Diethe.
\newblock Optimal continual learning has perfect memory and is np-hard.
\newblock In \emph{Proceedings of the 37th International Conference on Machine Learning}, ICML'20, 2020.

\bibitem[Koh et~al.(2020)Koh, Sagawa, Marklund, Xie, Zhang, Balsubramani, Hu, Yasunaga, Phillips, Gao, Lee, David, Stavness, Guo, Earnshaw, Haque, Beery, Leskovec, Kundaje, Pierson, Levine, Finn, and Liang]{koh2020wilds}
Pang~Wei Koh, Shiori Sagawa, Henrik Marklund, Sang~Michael Xie, Marvin Zhang, Akshay Balsubramani, Weihua Hu, Michihiro Yasunaga, Richard~Lanas Phillips, Irena Gao, Tony Lee, Etienne David, Ian Stavness, Wei Guo, Berton~A. Earnshaw, Imran~S. Haque, Sara Beery, Jure Leskovec, Anshul Kundaje, Emma Pierson, Sergey Levine, Chelsea Finn, and Percy Liang.
\newblock Wilds: A benchmark of in-the-wild distribution shifts, 2020.

\bibitem[Kothawade et~al.(2021)Kothawade, Beck, Killamsetty, and Iyer]{similar}
Suraj Kothawade, Nathan Beck, Krishnateja Killamsetty, and Rishabh Iyer.
\newblock {SIMILAR}: Submodular information measures based active learning in realistic scenarios.
\newblock \emph{Advances in Neural Information Processing Systems}, 34, 2021.

\bibitem[Kothawade et~al.(2022{\natexlab{a}})Kothawade, Chopra, Ghosh, and Iyer]{kothawade2022add}
Suraj Kothawade, Shivang Chopra, Saikat Ghosh, and Rishabh Iyer.
\newblock Active data discovery: Mining unknown data using submodular information measures, 2022{\natexlab{a}}.

\bibitem[Kothawade et~al.(2022{\natexlab{b}})Kothawade, Ghosh, Shekhar, Xiang, and Iyer]{talisman}
Suraj Kothawade, Saikat Ghosh, Sumit Shekhar, Yu~Xiang, and Rishabh~K. Iyer.
\newblock Talisman: Targeted active learning for object detection with rare classes and slices using submodular mutual information.
\newblock In \emph{Computer Vision - {ECCV} 2022 - 17th European Conference}, 2022{\natexlab{b}}.

\bibitem[Kothawade et~al.(2022{\natexlab{c}})Kothawade, Kaushal, Ramakrishnan, Bilmes, and Iyer]{prism}
Suraj Kothawade, Vishal Kaushal, Ganesh Ramakrishnan, Jeff~A. Bilmes, and Rishabh~K. Iyer.
\newblock {PRISM:} {A} rich class of parameterized submodular information measures for guided data subset selection.
\newblock In \emph{Thirty-Sixth {AAAI} Conference on Artificial Intelligence, {AAAI}}, pages 10238--10246, 2022{\natexlab{c}}.

\bibitem[Kothawade et~al.(2022{\natexlab{d}})Kothawade, Savarkar, Iyer, Ramakrishnan, and Iyer]{clinical}
Suraj Kothawade, Atharv Savarkar, Venkat Iyer, Ganesh Ramakrishnan, and Rishabh Iyer.
\newblock Clinical: Targeted active learning for imbalanced medical image classification.
\newblock In \emph{Medical Image Learning with Limited and Noisy Data: First International Workshop, MILLanD 2022, Held in Conjunction with MICCAI}, 2022{\natexlab{d}}.

\bibitem[Li et~al.(2022)Li, Kothawade, Chen, and Iyer]{PLATINUM}
Changbin Li, Suraj Kothawade, Feng Chen, and Rishabh Iyer.
\newblock {PLATINUM}: Semi-supervised model agnostic meta-learning using submodular mutual information.
\newblock In \emph{Proceedings of the 39th International Conference on Machine Learning}, volume 162 of \emph{Proceedings of Machine Learning Research}, pages 12826--12842. PMLR, 17--23 Jul 2022.

\bibitem[Li et~al.(2025)Li, Tan, Yang, Cheng, Dong, and Yang]{li2025adaptive}
Linhao Li, Yongzhang Tan, Siyuan Yang, Hao Cheng, Yongfeng Dong, and Liang Yang.
\newblock Adaptive decision boundary for few-shot class-incremental learning.
\newblock In \emph{Proceedings of the AAAI Conference on Artificial Intelligence}, volume~39, pages 18359--18367, 2025.

\bibitem[Lin and Bilmes(2011)]{submod_diversity}
Hui Lin and Jeff Bilmes.
\newblock A class of submodular functions for document summarization.
\newblock In \emph{Proceedings of the 49th Annual Meeting of the Association for Computational Linguistics: Human Language Technologies}, 2011.

\bibitem[Lin and Bilmes(2012)]{learningmixtures}
Hui Lin and Jeff~A. Bilmes.
\newblock Learning mixtures of submodular shells with application to document summarization, 2012.
\newblock URL \url{https://arxiv.org/abs/1210.4871}.

\bibitem[Lin et~al.(2014)Lin, Maire, Belongie, Hays, Perona, Ramanan, Doll{\'a}r, and Zitnick]{coco}
Tsung-Yi Lin, Michael Maire, Serge Belongie, James Hays, Pietro Perona, Deva Ramanan, Piotr Doll{\'a}r, and C.~Lawrence Zitnick.
\newblock Microsoft coco: Common objects in context.
\newblock In \emph{Computer Vision -- ECCV 2014}, Cham, 2014. Springer International Publishing.

\bibitem[Lin et~al.(2017)Lin, Goyal, Girshick, He, and Dollár]{RetinaNet}
Tsung-Yi Lin, Priya Goyal, Ross Girshick, Kaiming He, and Piotr Dollár.
\newblock Focal loss for dense object detection.
\newblock In \emph{2017 IEEE International Conference on Computer Vision (ICCV)}, pages 2999--3007, 2017.

\bibitem[Liu et~al.(2016)Liu, Anguelov, Erhan, Szegedy, Reed, Fu, and Berg]{SSD}
Wei Liu, Dragomir Anguelov, Dumitru Erhan, Christian Szegedy, Scott~E. Reed, Cheng-Yang Fu, and Alexander~C. Berg.
\newblock Ssd: Single shot multibox detector.
\newblock In \emph{ECCV (1)}, volume 9905, 2016.

\bibitem[Ma et~al.(2023{\natexlab{a}})Ma, Wang, Fan, yu~Wei, Li, Liu, and Lv]{Ma2023CATLA}
Shuailei Ma, Yuefeng Wang, Jiaqi Fan, Ying yu~Wei, Thomas~H. Li, Hongli Liu, and Fanbing Lv.
\newblock Cat: Localization and identification cascade detection transformer for open-world object detection.
\newblock In \emph{CVPR}, 2023{\natexlab{a}}.

\bibitem[Ma et~al.(2023{\natexlab{b}})Ma, Wang, Wei, Fan, Li, Liu, and Lv]{cat}
Shuailei Ma, Yuefeng Wang, Ying Wei, Jiaqi Fan, Thomas~H. Li, Hongli Liu, and Fanbing Lv.
\newblock Cat: Localization and identification cascade detection transformer for open-world object detection.
\newblock In \emph{2023 IEEE/CVF Conference on Computer Vision and Pattern Recognition (CVPR)}, pages 19681--19690, 2023{\natexlab{b}}.

\bibitem[Ma et~al.(2023{\natexlab{c}})Ma, Li, Zhang, Guo, Zhang, Gong, and Liu]{allow}
Yuqing Ma, Hainan Li, Zhange Zhang, Jinyang Guo, Shanghang Zhang, Ruihao Gong, and Xianglong Liu.
\newblock Annealing-based label-transfer learning for open world object detection.
\newblock In \emph{Proceedings of the IEEE/CVF Conference on Computer Vision and Pattern Recognition (CVPR)}, pages 11454--11463, 06 2023{\natexlab{c}}.

\bibitem[Maaz et~al.(2022)Maaz, Rasheed, Khan, Khan, Anwer, and Yang]{maaz2022multimodal}
Muhammad Maaz, Hanoona Rasheed, Salman Khan, Fahad~Shahbaz Khan, Rao~Muhammad Anwer, and Ming-Hsuan Yang.
\newblock Class-agnostic object detection with multi-modal transformer.
\newblock In \emph{European Conference on Computer Vision (ECCV)}. Springer, 2022.

\bibitem[Majee et~al.(2021)Majee, Agrawal, and Subramanian]{majee2021fewshot}
Anay Majee, Kshitij Agrawal, and Anbumani Subramanian.
\newblock {Few-Shot Learning For Road Object Detection}.
\newblock In \emph{AAAI Workshop on Meta-Learning and MetaDL Challenge}, volume 140, pages 115--126, 2021.

\bibitem[Majee et~al.(2024{\natexlab{a}})Majee, Kothawade, Killamsetty, and Iyer]{score}
Anay Majee, Suraj~Nandkishor Kothawade, Krishnateja Killamsetty, and Rishabh~K Iyer.
\newblock {SC}o{R}e: Submodular combinatorial representation learning.
\newblock In \emph{Proceedings of the 41st International Conference on Machine Learning}, volume 235, pages 34327--34349, 2024{\natexlab{a}}.

\bibitem[Majee et~al.(2024{\natexlab{b}})Majee, Sharp, and Iyer]{smile}
Anay Majee, Ryan Sharp, and Rishabh Iyer.
\newblock Smile: Leveraging submodular mutual information for robust few-shot object detection.
\newblock In \emph{European Conference on Computer Vision (ECCV)}, 2024{\natexlab{b}}.

\bibitem[Margatina et~al.(2021)Margatina, Vernikos, Barrault, and Aletras]{ALByContrastiveEx}
Katerina Margatina, Giorgos Vernikos, Loïc Barrault, and Nikolaos Aletras.
\newblock Active learning by acquiring contrastive examples.
\newblock \emph{arXiv preprint arXiv:2109.03764}, 2021.

\bibitem[Miller et~al.(2018)Miller, Nicholson, Dayoub, and S\"{u}nderhauf]{ose_ref}
Dimity Miller, Lachlan Nicholson, Feras Dayoub, and Niko S\"{u}nderhauf.
\newblock Dropout sampling for robust object detection in open-set conditions.
\newblock In \emph{2018 IEEE International Conference on Robotics and Automation (ICRA)}, 2018.

\bibitem[Mirzasoleiman et~al.(2015)Mirzasoleiman, Badanidiyuru, Karbasi, Vondrak, and Krause]{Mirzasoleiman2015lazierthanlazy}
Baharan Mirzasoleiman, Ashwinkumar Badanidiyuru, Amin Karbasi, Jan Vondrak, and Andreas Krause.
\newblock Lazier than lazy greedy.
\newblock \emph{Proceedings of the AAAI Conference on Artificial Intelligence}, 29\penalty0 (1), 2015.

\bibitem[Nemhauser et~al.(1978)Nemhauser, Wolsey, and Fisher]{Nemhauser1978}
G.~L. Nemhauser, L.~A. Wolsey, and M.~L. Fisher.
\newblock An analysis of approximations for maximizing submodular set functions---i.
\newblock \emph{Mathematical Programming}, 14\penalty0 (1):\penalty0 265--294, 1978.

\bibitem[Okanovic et~al.(2024)Okanovic, Waleffe, Mageirakos, Nikolakakis, Karbasi, Kalogerias, G{\"u}rel, and Rekatsinas]{r2s}
Patrik Okanovic, Roger Waleffe, Vasilis Mageirakos, Konstantinos Nikolakakis, Amin Karbasi, Dionysios Kalogerias, Nezihe~Merve G{\"u}rel, and Theodoros Rekatsinas.
\newblock Repeated random sampling for minimizing the time-to-accuracy of learning.
\newblock In \emph{The Twelfth International Conference on Learning Representations}, 2024.

\bibitem[Peng et~al.(2020)Peng, Zhao, and Lovell]{F_ILOD}
Can Peng, Kun Zhao, and Brian~C Lovell.
\newblock Faster ilod: Incremental learning for object detectors based on faster rcnn.
\newblock \emph{PRL}, 2020.

\bibitem[Redmon and Farhadi(2016)]{YOLOv2}
Joseph Redmon and Ali Farhadi.
\newblock Yolo9000: Better, faster, stronger.
\newblock \emph{2017 IEEE Conference on Computer Vision and Pattern Recognition (CVPR)}, pages 6517--6525, 2016.

\bibitem[Redmon et~al.(2015)Redmon, Divvala, Girshick, and Farhadi]{yolo}
Joseph Redmon, Santosh Divvala, Ross Girshick, and Ali Farhadi.
\newblock You only look once: Unified, real-time object detection, 2015.

\bibitem[Ren et~al.(2015)Ren, He, Girshick, and Sun]{frcnn}
Shaoqing Ren, Kaiming He, Ross Girshick, and Jian Sun.
\newblock Faster r-cnn: Towards real-time object detection with region proposal networks.
\newblock In \emph{Advances in Neural Information Processing Systems}, volume~28. Curran Associates, Inc., 2015.

\bibitem[Sener and Savarese(2018)]{sener2018active}
Ozan Sener and Silvio Savarese.
\newblock Active learning for convolutional neural networks: A core-set approach.
\newblock In \emph{ICLR}, 2018.

\bibitem[Shmelkov et~al.(2017)Shmelkov, Schmid, and Alahari]{ILOD}
Konstantin Shmelkov, Cordelia Schmid, and Karteek Alahari.
\newblock Incremental learning of object detectors without catastrophic forgetting.
\newblock In \emph{ICCV}, 2017.

\bibitem[Sun et~al.(2024)Sun, Li, and Mu]{sun2024exploring}
Zhicheng Sun, Jinghan Li, and Yadong Mu.
\newblock Exploring orthogonality in open world object detection.
\newblock In \emph{Proceedings of the IEEE/CVF Conference on Computer Vision and Pattern Recognition}, pages 17302--17312, 2024.

\bibitem[Vaswani et~al.(2017)Vaswani, Shazeer, Parmar, Uszkoreit, Jones, Gomez, Kaiser, and Polosukhin]{vaswani2017attention}
Ashish Vaswani, Noam Shazeer, Niki Parmar, Jakob Uszkoreit, Llion Jones, Aidan~N Gomez, \L~ukasz Kaiser, and Illia Polosukhin.
\newblock Attention is all you need.
\newblock In \emph{Advances in Neural Information Processing Systems}, volume~30, 2017.

\bibitem[Wang et~al.(2024{\natexlab{a}})Wang, Vaze, and Han]{wang2024dissect}
Hongjun Wang, Sagar Vaze, and Kai Han.
\newblock Dissecting out-of-distribution detection and open-set recognition: A critical analysis of methods and benchmarks.
\newblock \emph{International Journal of Computer Vision (IJCV)}, 2024{\natexlab{a}}.

\bibitem[Wang et~al.(2023)Wang, Yue, Hua, and Zhang]{Randbox}
Yanghao Wang, Zhongqi Yue, Xian-Sheng Hua, and Hanwang Zhang.
\newblock Random boxes are open-world object detectors.
\newblock In \emph{Proceedings of the IEEE/CVF International Conference on Computer Vision (ICCV)}, pages 6233--6243, October 2023.

\bibitem[Wang et~al.(2024{\natexlab{b}})Wang, Mu, Zhu, and Hu]{wang2024os}
Yu~Wang, Junxian Mu, Pengfei Zhu, and Qinghua Hu.
\newblock Exploring diverse representations for open set recognition.
\newblock AAAI, 2024{\natexlab{b}}.

\bibitem[Wei et~al.(2014)Wei, Liu, Kirchhoff, Bartels, and Bilmes]{speech_subset_selection}
Kai Wei, Yuzong Liu, Katrin Kirchhoff, Chris Bartels, and Jeff Bilmes.
\newblock Submodular subset selection for large-scale speech training data.
\newblock In \emph{2014 IEEE International Conference on Acoustics, Speech and Signal Processing (ICASSP)}, 2014.
\newblock \doi{10.1109/ICASSP.2014.6854213}.

\bibitem[Wei et~al.(2015)Wei, Iyer, and Bilmes]{wei15_subset}
Kai Wei, Rishabh Iyer, and Jeff Bilmes.
\newblock Submodularity in data subset selection and active learning.
\newblock In \emph{ICML}, 2015.

\bibitem[Wu et~al.(2022{\natexlab{a}})Wu, Zhao, Ma, Wang, and Liu]{wu2022two}
Yan Wu, Xiaowei Zhao, Yuqing Ma, Duorui Wang, and Xianglong Liu.
\newblock Two-branch objectness-centric open world detection.
\newblock In \emph{Proceedings of the 3rd International Workshop on Human-Centric Multimedia Analysis}, pages 35--40. ACM, 2022{\natexlab{a}}.

\bibitem[Wu et~al.(2022{\natexlab{b}})Wu, Lu, Chen, Wu, Kang, and Yu]{wu2022ucowod}
Zhiheng Wu, Yue Lu, Xingyu Chen, Zhengxing Wu, Liwen Kang, and Junzhi Yu.
\newblock Uc-owod: Unknown-classified open world object detection.
\newblock \emph{arXiv preprint arXiv:2203.09451}, 2022{\natexlab{b}}.

\bibitem[Xi et~al.(2024)Xi, Huang, Zhong, and Luo]{xi2024umb}
Xing Xi, Yangyang Huang, Zhijie Zhong, and Ronghua Luo.
\newblock {UMB}: Understanding model behavior for open-world object detection.
\newblock In \emph{The Thirty-eighth Annual Conference on Neural Information Processing Systems}, 2024.

\bibitem[Yang et~al.(2022{\natexlab{a}})Yang, Deng, Shi, Li, Zhang, Xu, Zhao, Lin, and Liang]{ROSETTA}
Binbin Yang, Xi~Deng, Han Shi, Changlin Li, Gengwei Zhang, Hang Xu, Shen Zhao, Liang Lin, and Xiaodan Liang.
\newblock Continual object detection via prototypical task correlation guided gating mechanism.
\newblock \emph{2022 IEEE/CVF Conference on Computer Vision and Pattern Recognition (CVPR)}, pages 9245--9254, 2022{\natexlab{a}}.

\bibitem[Yang et~al.(2022{\natexlab{b}})Yang, Sun, Jiang, Xia, Zhang, Yuan, Wang, Luo, and Xu]{OST}
Shuo Yang, Peize Sun, Yi~Jiang, Xiaobo Xia, Ruiheng Zhang, Zehuan Yuan, Changhu Wang, Ping Luo, and Min Xu.
\newblock Objects in semantic topology.
\newblock In \emph{The Tenth International Conference on Learning Representations, ICLR 2022, Virtual Event, April 25-29, 2022}, 2022{\natexlab{b}}.

\bibitem[Yu et~al.(2022)Yu, Ma, Li, Peng, and Xie]{yu2022open}
Jinan Yu, Liyan Ma, Zhenglin Li, Yan Peng, and Shaorong Xie.
\newblock Open-world object detection via discriminative class prototype learning.
\newblock In \emph{2022 IEEE International Conference on Image Processing (ICIP)}, pages 626--630. IEEE, 2022.

\bibitem[Zhang et~al.(2024)Zhang, Chen, Zou, Li, and Li]{zhang2024learning}
Zhenyu Zhang, Guangyao Chen, Yixiong Zou, Yuhua Li, and Ruixuan Li.
\newblock Learning unknowns from unknowns: Diversified negative prototypes generator for few-shot open-set recognition.
\newblock In \emph{Proceedings of the 32nd ACM International Conference on Multimedia}, pages 6053--6062, 2024.

\bibitem[Zhu et~al.(2020)Zhu, Su, Lu, Li, Wang, and Dai]{zhu2020deformable}
Xizhou Zhu, Weijie Su, Lewei Lu, Bin Li, Xiaogang Wang, and Jifeng Dai.
\newblock Deformable detr: Deformable transformers for end-to-end object detection.
\newblock \emph{arXiv preprint arXiv:2010.04159}, 2020.

\bibitem[Zhu et~al.(2021)Zhu, Su, Lu, Li, Wang, and Dai]{D-DETR}
Xizhou Zhu, Weijie Su, Lewei Lu, Bin Li, Xiaogang Wang, and Jifeng Dai.
\newblock Deformable {\{}detr{\}}: Deformable transformers for end-to-end object detection.
\newblock In \emph{International Conference on Learning Representations}, 2021.

\bibitem[Zohar et~al.(2023)Zohar, Wang, and Yeung]{Zohar_2023_CVPR}
Orr Zohar, Kuan-Chieh Wang, and Serena Yeung.
\newblock Prob: Probabilistic objectness for open world object detection.
\newblock In \emph{Proceedings of the IEEE/CVF Conference on Computer Vision and Pattern Recognition (CVPR)}, pages 11444--11453, June 2023.

\end{thebibliography}
}


\newpage
\appendix

\section{Appendix}

\startcontents[sections]
\printcontents[sections]{l}{1}{\setcounter{tocdepth}{3}}

\subsection{Notation}
Following the problem definition in the main paper we introduce the notations used in \cref{tab:notations} throughout the paper.
\begin{table*}[th]
      \caption{Collection of notations used in the paper.}
      \centering
      \resizebox{\textwidth}{!}{
        \begin{tabular}{ c | c }
            \toprule
           \textbf{Symbol}  & \textbf{Description} \\
            \midrule
            $t$ & Task identifier for each OWOD task. \\
            $T_t$ & Each Task in OWOD. \\
            $D^t$ & Training dataset for each task. \\
            $\mathcal{T}$ & The Ground set, here refers to the mini-batch at each iteration. \\
            $K^t$ & Complete set of currently known classes in $T_t$. \\
            $K^{t-1}$ & Complete set of previously known classes in $T_t$. \\
            $U^t$ & Complete set of unknown classes in $T_t$. \\
            $\hat{K}^t$ & Predefined Replay buffer of currently known classes in $T_t$. \\
            $\hat{K}^{t-1}$ & Predefined Replay buffer of previously known classes in $T_t$. \\
            $h^t(x, \theta)$ & Task specific Object Detector used as feature extractor. \\
            $Clf(.,.)$ & Multi-Layer Perceptron as classifier. In our case a two layer network. \\
            $\theta$ & Parameters of the feature extractor. \\
            $s_{A,B}(\theta)$ & Cross-Similarity between sets $A, B \in \mathcal{V}$. \\
            $s_A(\theta)$ & Self-Similarity between samples in set $A \in \mathcal{T}$. \\
            $f(A)$ & Submodular Information function over a set $A$. \\
            $H_f(A|Q)$ & Submodular Conditional Gain function between sets $A$ and $Q$. \\
            $L_{\text{CROWD}}(\theta)$ & Loss value computed over all known and unknown objects. \\
            $L_{\text{CROWD}}^{self}(\theta)$ & Combinatorial loss computed over all known classes $K^t_i \in \mathcal{T}$. \\
            $L_{\text{CROWD}}^{cross}(\theta)$ & Combinatorial loss computed between known classes $K^t_i$ and unknown classes $U^t$. \\
            \bottomrule
      \end{tabular}
     }
     \label{tab:notations}      
\end{table*}

\subsection{Additional Related Work}
\label{app:additional_rel_work}
\noindent \textbf{Data subset selection} aims at identifying a distinct set of examples from a large pool which accurately captures the properties of the data distribution. 
This has rendered subset selection to be a natural choice for data-efficient machine learning tasks like Active Learning~\cite{similar,ALByContrastiveEx,sener2018active,ducoffe2018adversarial}, Continual Learning~\cite{OptimalContinualLearning,Gradient_based_sample_selection}, Data Summarization~\cite{similar,prism} etc. 
Traditionally subset selection has been defined as a subsampling technique based on similarity~\cite{orient}, uncertainty~\cite{SelectionviaProxy} etc. or random~\cite{r2s}.
Orthogonally, a new line of work based on combinatorial functions, particularly submodular functions~\cite{fujishige, alt_gen_submod} have emerged which effectively selects informative subsets by modeling the notions of cooperation, diversity and representation~\cite{prism}. These functions formulate subset selection as a greedy maximization task~\cite{Mirzasoleiman2015lazierthanlazy} based on several information theoretic measures like Total Information, Mutual Information, Conditional Gain etc. (discussed in \cref{sec:submod_prelims} of the main paper). 
Concurrent to their success in vision~\cite{prism,similar}, language~\cite{learningmixtures}, speech~\cite{speech_subset_selection} etc. domains, subset selection has been used in auxiliary learning mechanisms like meta-learning~\cite{PLATINUM} and data-discovery~\cite{kothawade2022add} targeting identification of rare or unseen examples from an unlabeled example pool. CROWD exploits this line of investigation adopting a combinatorial subset selection technique (detailed in \cref{sec:crowd_d} to discover unknown objects in the open-world setting).\looseness-1

\noindent \textbf{Object detection} (OD) is a fundamental task in computer vision encapsulating both localization and recognition tasks under the same roof. 
OD methods are traditionally grouped into two principal paradigms: single-stage and two-stage detectors. Single-stage detectors, exemplified by SSD~\cite{SSD}, RetinaNet~\cite{RetinaNet}, and YOLO~\cite{yolo,YOLOv2}, CenterNet~\cite{centernet} unify the processes of object localization and classification into a single feed-forward network, enabling real-time performance with relatively low computational overhead. In contrast, two-stage detectors, such as Faster R-CNN~\cite{rcnn, FastRCNN, frcnn}, adopt a cascaded architecture wherein a Region Proposal Network (RPN) first hypothesizes candidate object regions, followed by a refinement stage that simultaneously predicts the class and precise bounding box of each proposal. 
CNN based architectures struggles with the long-range dependencies, which is important for understanding the complex spatial relationships between objects at varying scales (perspective views). Transformer based models~\cite{DETR,D-DETR,UP-DETR} improve upon this vulnerability by introducing a self-attention~\cite{vaswani2017attention} mechanism based on an encoder-decoder architecture~\cite{DETR}. While these models achieve impressive performance in closed-world settings (all object categories present during testing are known and predefined in the training data) they under-perform in open-world scenarios when encountering unknown objects unseen during training.\looseness-1

\noindent \textbf{Preliminaries of Submodularity} (continued from \cref{sec:submod_prelims}) As discussed in \cref{sec:submod_prelims} of the main paper, submodular functions have been recognized to model notions of cooperation~\cite{submod_cooperation}, diversity~\cite{submod_diversity}, representation~\cite{prism} and coverage~\cite{vid_sum_2019}. 
Following the combinatorial formulation in \cref{sec:prob_def} of the main paper we define the ground set $\mathcal{V} = \{A_1, A_2, \cdots A_N\}$, s.t. $|\mathcal{V}| = N$ and explore four different categories of submodular information functions in our work, namely - 

\noindent \textbf{(1)} \textit{Submodular Total Information} ($S_f$) which measures the total information contained in each set~\cite{fujishige}, expressed as $S_f(A_1, A_2, \dots, A_N)$ as in \cref{eq:sim}. Maximizing $S_f$ over a set $A_i$ models diversity~\cite{submod_diversity} while minimizing $S_f$ models cooperation~\cite{submod_cooperation}.\looseness-1
\begin{align}
    S_f(A_1, A_2, \dots, A_N) = \overset{N}{\underset{i=1}{\sum}}f(A_i)
    \label{eq:sim}
\end{align}


\noindent \textbf{(2)} \textit{Submodular Conditional Gain} ($H_f$) which models the gain in information when a set $A_j$ is added to $A_i$. $H_f$ models the notion of \textit{dissimilarity} between sets and can be expressed in \cref{eq:scg}.
\begin{align}
\begin{split}
    H_f(A_i| A_j) &= f(A_i \cup A_j) - f(A_j) \text{ , }  \forall i,j \in |\mathcal{V}| \\ 
\end{split}
\label{eq:scg}    
\end{align}


\begin{wrapfigure}{r}{0.5\textwidth}
\begin{minipage}{0.5\textwidth}
\begin{algorithm}[H]
\caption{Greedy Submodular Maximization~\cite{Nemhauser1978}}
\label{alg:greedy}
\begin{algorithmic}[1]
\Require Submodular function $f : 2^{\mathcal{V}} \rightarrow \mathbb{R}$, cardinality constraint $\mathtt{k}$
\Ensure Set $A \subseteq \mathcal{V}$ maximizing $f(A)$ under cardinality constraint $\mathtt{k}$
\State $A \leftarrow \emptyset$
\For{$j = 1$ to $\mathtt{k}$}
    \State $e \leftarrow \arg\max\limits_{v \in \mathcal{V} \setminus A} \left[f(A \cup \{v\}) - f(A)\right]$
    \State $A \leftarrow A \cup \{e\}$
\EndFor
\State \Return $A$
\end{algorithmic}
\end{algorithm}
\end{minipage}
\end{wrapfigure}
 
Given a submodular function $f$ (can alternatively be $H_f$) tasks like selection~\cite{jain2023efficient,killamsetty_automata} and summarization~\cite{kaushal2019framework, vid_sum_2019} have been modeled as a discrete optimization problem to identify a summarized set of examples $A \subseteq \mathcal{V}$ via submodular maximization under a cardinality constraint ($|A| \leq k$), i.e. $\max_{A \subseteq \mathcal{V}, |A| \leq k} f(A)$. This can be fairly approximated with a $(1 - e^{-1})$ constant factor guarantee~\cite{Nemhauser1978} using greedy optimization techniques~\cite{Mirzasoleiman2015lazierthanlazy} as shown in \cref{alg:greedy}.
Extending the definition of submodular functions to continuous optimization space \citet{score} have proposed a set of novel family of learning objectives which minimize total information and total correlation among sets in $D_{train}$ using continuous optimization techniques like SGD. These objectives have been shown to be significantly more robust to large imbalance demonstrated in real-world tasks like longtail recognition~\cite{score} and few-shot learning~\cite{smile}.\looseness-1

\subsection{Derivations of Instances of $L_{\text{CROWD}}$}
\label{app:crowd_derivations}
As discussed in \cref{sec:crowd_l} of the main paper, varying the choice of Submodular function $f$ in \cref{eq:crowd_loss} results in several instances of $L_{\text{CROWD}}$. Based on three popular choices of $f$ among Facility-Location, Graph-Cut and Log-Determinant, we derive the respective formulations of $L_{\text{CROWD}}$. Note, that the derivations of $L_{\text{CROWD}}^{self}$ are adapted from \citet{score} and are thus not included below.

\subsubsection{Derivation of CROWD-FL}
\begin{theorem}
    Given a set of known RoIs $K^t_i$, $i \in [1, C^t]$, a set of unknown RoIs $U^t$ ($\mathcal{T} = K^t \cup U^t$) and the Facility-Location based submodular function $f$ defined over any set $A$ s.t. $f(A) = \sum_{i \in \mathcal{T}} \max_{j \in A} s_{ij}$, we define CROWD-FL learning objective to learn the parameters $\theta$ of the model $h^t$, containing two components $L_{\text{CROWD}}^{self}$ and $L_{CROWD}^{cross}$ as shown in \cref{eq:crowd_fl}. Here, $s_ij$ resembles the similarity between samples $i$ and $j$ respectively.
\end{theorem}

\begin{align}
\begin{split}
    L_{CROWD}^{self} &= \sum_{i = 1}^{C^t} \frac{1}{|K^t_i|}\sum_{i \in \mathcal{T} \setminus K_i^t} \max_{j \in K_i^t} s_{ij}(\theta) \\
    L_{CROWD}^{cross}(\theta) &= \sum_{i = 1}^{C^t} \frac{1}{|\mathcal{T}|}\sum\limits_{n \in \mathcal{T}} \max(\max\limits_{k \in K_i^t} s_{nk}(\theta)-\nu \max\limits_{u \in U^t} s_{nu}(\theta), 0)
\end{split}
\label{eq:crowd_fl}
\end{align}

\begin{proof}
    From the definition of $L_{\text{CROWD}}^{cross}$ in \cref{eq:crowd_loss} we find,
    \begin{align}
        \begin{split}
            L_{\text{CROWD}}^{cross} &= \sum_{i = 1}^{C^t} H_f(K^t_i | U^t) \\
            L_{\text{CROWD}}^{cross} &= \sum_{i = 1}^{C^t} f(K^t_i \cup U^t) - f(U^t)            
        \end{split}
    \end{align}
    Substituting the definition of $f(A)$ over any set from the theorem in the above expression we get -
    \begin{align}
        \begin{split}
            L_{\text{CROWD}}^{cross} &= \sum_{i = 1}^{C^t} H_f(K^t_i | U^t) \\
            L_{\text{CROWD}}^{cross} &= \sum_{i = 1}^{C^t} \sum_{n \in \mathcal{T}} \underset{k \in K^t_i \cup U^t}{\max} s_{nk} - \sum_{n \in \mathcal{T}} \underset{u \in U^t}{\max} s_{nu} \\
            L_{\text{CROWD}}^{cross} &= \sum_{i = 1}^{C^t} \sum_{n \in \mathcal{T}} \max \Biggl(\underset{k \in K^t_i}{\max} s_{nk}, \underset{k U^t}{\max} s_{nk}\Biggr) - \sum_{n \in \mathcal{T}} \underset{u \in U^t}{\max} s_{nu} \\
            L_{\text{CROWD}}^{cross} &= \sum_{i = 1}^{C^t} \sum_{n \in \mathcal{T}} \max \Biggl(\underbrace{\underset{k \in K^t_i}{\max} s_{nk}}_{\text{Term 1}} - \underbrace{\underset{u \in U^t}{\max} s_{nu}}_{\text{Term 2}}, 0\Biggr)
        \end{split}
    \end{align}
    The Term 2 in the above equation controls the degree of separation between $K^t_i$ and $U^t$. Due to this we introduce a hyper-parameter $\nu$ which we can control during model training. Sine $\nu$ is a constant it does not affect the submodular properties of $L_{\text{CROWD}}^{cross}$. The final loss formulation, normalized by the size of $\mathcal{T}$ thus becomes -
    \begin{align}
        L_{\text{CROWD}}^{cross} &= \sum_{i = 1}^{C^t} \frac{1}{|\mathcal{T}|} \sum_{n \in \mathcal{T}} \max \Biggl(\underset{k \in K^t_i}{\max} s_{nk} - \nu \underset{u \in U^t}{\max} s_{nu}, 0\Biggr)
    \end{align}

    Additionally, we do not provide proofs for $L_{\text{CROWD}}^{self}$ since this function largely resembles the total information formulation in \citet{score}.
\end{proof}
\subsubsection{Derivation of CROWD-GC}
\begin{theorem}
    Given a set of known RoIs $K^t_i$, $i \in [1, C^t]$, a set of unknown RoIs $U^t$ ($\mathcal{T} = K^t \cup U^t$) and the Graph-Cut based submodular function $f$ defined over any set $A$ s.t. $f(A) = \sum_{i \in \mathcal{T}} \sum_{j \in A} s_{ij} - \lambda \sum_{i,j \in A} s_{ij}$, we define CROWD-GC learning objective to learn the parameters $\theta$ of the model $h^t$ containing two components $L_{\text{CROWD}}^{self} (\theta)$ and $L_{CROWD}^{cross}(\theta)$ as shown in \cref{eq:crowd_gc}. Here, $s_ij$ resembles the similarity between samples $i$ and $j$ respectively.
\end{theorem}

\begin{align}
\begin{split}
    L_{CROWD}^{self} &= \sum_{i = 1}^{C^t} \frac{1}{|K_i^t|}[\sum_{i \in K_i^t}\sum_{j \in \mathcal{T} \setminus U^t}s_{ij}(\theta) - \lambda \sum_{i, j \in K_i^t} s_{ij}(\theta)] \\
    L_{CROWD}^{cross}(\theta) &= \sum_{i = 1}^{C^t} \frac{1}{|\mathcal{T}|} [f(K_i^t; \theta) - 2 \lambda \nu \sum\limits_{k \in K_i^t, u \in U_i^t} s_{ku}(\theta)]
\end{split}
\label{eq:crowd_gc}
\end{align}

\begin{proof}
    From the definition of $L_{\text{CROWD}}^{cross}$ in \cref{eq:crowd_loss} we find,
    \begin{align}
        \begin{split}
            L_{\text{CROWD}}^{cross} &= \sum_{i = 1}^{C^t} H_f(K^t_i | U^t) \\
            L_{\text{CROWD}}^{cross} &= \sum_{i = 1}^{C^t} f(K^t_i \cup U^t) - f(U^t)            
        \end{split}
    \end{align}
    Substituting the definition of $f(A)$ over any set from the theorem in the above expression of $L_{\text{CROWD}}^{cross}$we get -
    \begin{align}
        \begin{split}
            L_{\text{CROWD}}^{cross} &= \sum_{i = 1}^{C^t} H_f(K^t_i | U^t) = \sum_{i = 1}^{C^t} f(K^t_i \cup U^t) - f(U^t)\\
            L_{\text{CROWD}}^{cross} &= \sum_{i = 1}^{C^t} \sum_{n \in \mathcal{T}} \sum_{k \in K^t_i \cup U^t} s_{nk} - \lambda \sum_{n, k \in K^t_i \cup U^t} s_{nk} - \sum_{n \in \mathcal{T}} \sum_{u \in U^t} s_{nu}  + \lambda \sum_{n, u \in U^t} s_{nu}  \\
            L_{\text{CROWD}}^{cross} &= \sum_{i = 1}^{C^t} \sum_{n \in \mathcal{T}} \sum_{k \in K^t_i} s_{nk} + \sum_{n \in \mathcal{T}} \sum_{u \in U^t} s_{nu} - \lambda \sum_{n, k \in K^t_i \cup U^t} s_{nk} \\
            &- \sum_{n \in \mathcal{T}} \sum_{u \in U^t} s_{nu} + \lambda \sum_{n, u \in U^t} s_{nu}
        \end{split}
    \end{align}
    The second term and the fourth term cancels out (same value with opposite signs).
    \begin{align}
        \begin{split}
            L_{\text{CROWD}}^{cross} &= \sum_{i = 1}^{C^t} \sum_{n \in \mathcal{T}} \sum_{k \in K^t_i} s_{nk} - \lambda \Biggl(\sum_{n, k \in K^t_i \cup U^t} s_{nk} + \sum_{n, u \in U^t} s_{nu}\Biggr) \\
            L_{\text{CROWD}}^{cross} &= \sum_{i = 1}^{C^t} \sum_{n \in \mathcal{T}} \sum_{k \in K^t_i} s_{nk} - \lambda \Biggl(\sum_{n, k \in K^t_i} s_{nk} + 2\sum_{n, u \in U^t} s_{nu}\Biggr) \\
        \end{split}
    \end{align}
    Now, rearranging the terms of the equation we get -
    \begin{align}
        \begin{split}
            L_{\text{CROWD}}^{cross} &= \sum_{i = 1}^{C^t} \Biggl(\sum_{n \in \mathcal{T}} \sum_{k \in K^t_i} s_{nk} - \lambda \sum_{n, k \in K^t_i} s_{nk}\Biggr) + 2\lambda\sum_{n, u \in U^t} s_{nu} \\
            L_{\text{CROWD}}^{cross} &= \sum_{i = 1}^{C^t} \underbrace{f(K^t_i)}_{\text{Term 1}} + \underbrace{2\lambda\sum_{n, u \in U^t} s_{nu}}_{\text{Term 2}} \\
        \end{split}
    \end{align}
    Similar to CROWD-FL the Term 2 in the above equation controls the degree of separation between $K^t_i$ and $U^t$. Due to this we introduce a hyper-parameter $\nu$ which we can control during model training. Sine $\nu$ is a constant it does not affect the submodular properties of $L_{\text{CROWD}}^{cross}$. The final loss formulation, normalized by the size of $\mathcal{T}$ thus becomes -
    \begin{align}
        \begin{split}
            L_{\text{CROWD}}^{cross} &= \sum_{i = 1}^{C^t} f(K^t_i) + 2\lambda\nu\sum_{n, u \in U^t} s_{nu} \\
        \end{split}
    \end{align}
    Additionally, we do not provide proofs for $L_{\text{CROWD}}^{self}$ since this function largely resembles the total information formulation in \citet{score}.
\end{proof}

\subsubsection{Derivation of CROWD-LogDet}
\begin{theorem}
    Given a set of known RoIs $K^t_i$, $i \in [1, C^t]$, a set of unknown RoIs $U^t$ ($\mathcal{T} = K^t \cup U^t$) and the Log-Determinant based submodular function $f$ defined over any set $A$ s.t. $f(A) = \log\det(s_A)$, we define CROWD-LogDet learning objective which contains two components $L_{\text{CROWD}}^{self}$ and $L_{CROWD}^{cross}$ as shown in \cref{eq:crowd_ld}. Here, $s_ij$ resembles the similarity between samples $i$ and $j$ respectively.
\end{theorem}

\begin{align}
\begin{split}
    L_{CROWD}^{self} &= \sum_{i = 1}^{C^t} \frac{1}{|K_i^t|}\log \det (s_{K_i^t}(\theta) + \lambda \mathbb{I}_{|K_i^t|}) \\
    L_{CROWD}^{cross}(\theta) &= \sum_{i = 1}^{C^t} \frac{1}{|\mathcal{T}|} \log\det(s_{K_i^t}(\theta) - \nu^2 s_{K_i^t, U^t}(\theta)s_{U^t}^{-1}(\theta)s_{K_i^t, U^t}(\theta)^T)
\end{split}
\label{eq:crowd_ld}
\end{align}

\begin{proof}
     From the definition of $L_{\text{CROWD}}^{cross}$ in \cref{eq:crowd_loss} we find,
    \begin{align}
        \begin{split}
            L_{\text{CROWD}}^{cross} &= \sum_{i = 1}^{C^t} H_f(K^t_i | U^t) \\
            L_{\text{CROWD}}^{cross} &= \sum_{i = 1}^{C^t} f(K^t_i \cup U^t) - f(U^t)            
        \end{split}
    \end{align}
    Substituting the definition of $f(A)$ over any set from the theorem in the above expression of $L_{\text{CROWD}}^{cross}$we get -
    \begin{align}
        \begin{split}
            L_{\text{CROWD}}^{cross} &= \sum_{i = 1}^{C^t} H_f(K^t_i | U^t) = \sum_{i = 1}^{C^t} f(K^t_i \cup U^t) - f(U^t)\\
            L_{\text{CROWD}}^{cross} &= \sum_{i = 1}^{C^t} \log \det (s_{K^t_i \cup U^t}) - \log \det (s_{U^t})\\
            &= \sum_{i = 1}^{C^t} \log \frac{\det (s_{K^t_i \cup U^t})}{\det (s_{U^t})}\\
        \end{split}
    \end{align}
    From Schur's complement which states that given two sets A and B $\det(s_{A \cup B}) = \det(s_A).\det(s_{A \cup B} \setminus s_A)$. Replacing the term $\det(s_{K^t_i \cup U^t})$ with the above definition we get -
    \begin{align}
        \begin{split}
            L_{\text{CROWD}}^{cross} &= \sum_{i = 1}^{C^t} \log \frac{\det(s_{U^t}) . \det (s_{K^t_i \cup U^t} \setminus s_{U^t})}{\det (s_{U^t})}\\
            &= \sum_{i = 1}^{C^t} \log \det (s_{K^t_i \cup U^t} \setminus s_{U^t})
        \end{split}
    \end{align}
    Following Schur's complement yet again which states that $s_{A \cup B} \setminus s_A = s_B - s_{A,B}^T s_A^{-1}s_{A,B}$, where $s_{A,B}$ refers to the cross-similarities between sets $A$ and $B$ while $s_A$ and $s_B$ represent the corresponding self-similarities and substitute this definition into the aforementioned equation as -
    \begin{align}
        \begin{split}
            L_{\text{CROWD}}^{cross} &= \sum_{i = 1}^{C^t} \log \det(s_{K^t_i} - s_{K^t_i,U^t} s_{U^t}^{-1}s_{K^t_i,U^t}^T) 
        \end{split}
    \end{align}
    Normalizing this term with the size of the ground set $|\mathcal{T}|$ and introducing the hyper-parameter $\nu$ which trades-off between inter-cluster separation and intra-cluster compactness, we derive the function for $L_{\text{CROWD}}^{cross}$ as -
    \begin{align}
        \begin{split}
            L_{\text{CROWD}}^{cross} &= \sum_{i = 1}^{C^t} \frac{1}{|\mathcal{T}|} \log \det(s_{K^t_i} - \nu^2s_{K^t_i,U^t} s_{U^t}^{-1}s_{K^t_i,U^t}^T) 
        \end{split}
    \end{align}
    Similar to previously derived objectives, we do not provide proofs for $L_{\text{CROWD}}^{self}$ since this function largely resembles the total information formulation in \citet{score}.
\end{proof}

\begin{table*}[t]
\centering
\caption{\textbf{Generalization performance on Incremental Object Detection (IOD)} where we show that our CROWD approach (here only CROWD-L) when applied to the fintetuning stage of IOD tasks show better generalizability. Best results are in \textbf{bold} while new classes introduced in the task are shaded \colorbox{lightgray}{gray}.}
\label{tab:iod_benchmark}
\setlength{\tabcolsep}{3pt}
\resizebox{\textwidth}{!}{%
\begin{tabular}{@{}lccccccccccccccccccccc@{}}
\toprule
{\textbf{10 + 10 setting}} & aero & cycle & bird & boat & bottle & bus & car & cat & chair & cow & table & dog & horse & bike & person & plant & sheep & sofa & train & tv & mAP \\ \midrule
ILOD \cite{ILOD} & 69.9 & 70.4 & 69.4 & 54.3 & 48 & 68.7 & 78.9 & 68.4 & 45.5 & 58.1 & \cellcolor{lightgray} 59.7 & \cellcolor{lightgray} 72.7 & \cellcolor{lightgray} 73.5 & \cellcolor{lightgray} 73.2 & \cellcolor{lightgray} 66.3 & \cellcolor{lightgray} 29.5 & \cellcolor{lightgray} 63.4 & \cellcolor{lightgray} 61.6 & \cellcolor{lightgray} 69.3 & \cellcolor{lightgray} 62.2 & 63.2 \\

Faster ILOD \cite{F_ILOD} & 72.8 & 75.7 & 71.2 & 60.5 & 61.7 & 70.4 & 83.3 & 76.6 & 53.1 & 72.3 & \cellcolor{lightgray} 36.7 & \cellcolor{lightgray} 70.9 & \cellcolor{lightgray} 66.8 & \cellcolor{lightgray} 67.6 & \cellcolor{lightgray} 66.1 & \cellcolor{lightgray} 24.7 & \cellcolor{lightgray} 63.1 & \cellcolor{lightgray} 48.1 & \cellcolor{lightgray} 57.1 & \cellcolor{lightgray} 43.6 & 62.1 \\ 
ORE ~\cite{joseph2021ore} & 63.5 & 70.9 & 58.9 & 42.9 & 34.1 & 76.2 & 80.7 & 76.3 & 34.1 & 66.1 & \cellcolor{lightgray} 56.1 & \cellcolor{lightgray} 70.4 & \cellcolor{lightgray} 80.2 & \cellcolor{lightgray} 72.3 & \cellcolor{lightgray} 81.8 & \cellcolor{lightgray} 42.7 & \cellcolor{lightgray} 71.6 & \cellcolor{lightgray} 68.1 & \cellcolor{lightgray} 77.0 & \cellcolor{lightgray} 67.7 & 64.5 \\
Meta-ILOD \cite{Meta_ILOD} & 76.0 & 74.6 & 67.5 & 55.9 & 57.6 & 75.1 & 85.4 &77.0 &43.7 &70.8 & \cellcolor{lightgray}  60.1 & \cellcolor{lightgray}  66.4 & \cellcolor{lightgray}  76.0 & \cellcolor{lightgray}  72.6 & \cellcolor{lightgray}  74.6 & \cellcolor{lightgray}  39.7 & \cellcolor{lightgray}  64.0 & \cellcolor{lightgray}  60.2 & \cellcolor{lightgray} 68.5 & \cellcolor{lightgray}  60.7 & 66.3 \\

ROSETTA \cite{ROSETTA} &74.2 & 76.2 & 64.9 & 54.4 &  57.4 & 76.1 & 84.4 & 68.8 & 52.4 &  67.0 & \cellcolor{lightgray}  62.9 & \cellcolor{lightgray}   63.3 & \cellcolor{lightgray}   79.8 &\cellcolor{lightgray}  72.8 & \cellcolor{lightgray}  78.1 & \cellcolor{lightgray}  40.1 &\cellcolor{lightgray}  62.3 & \cellcolor{lightgray}  61.2 & \cellcolor{lightgray}  72.4 & \cellcolor{lightgray}  66.8 & 66.8 \\

OW-DETR\cite{gupta2021ow} &61.8 & 69.1 & 67.8 & 45.8 & 47.3 & 78.3 &78.4 &78.6 &36.2 & 71.5 & \cellcolor{lightgray}  57.5 & \cellcolor{lightgray}  75.3 & \cellcolor{lightgray}  76.2 &\cellcolor{lightgray}  77.4 &\cellcolor{lightgray}  79.5 &\cellcolor{lightgray}  40.1 &\cellcolor{lightgray}  66.8 &\cellcolor{lightgray}  66.3 &\cellcolor{lightgray}  75.6 &\cellcolor{lightgray}  64.1 & 65.7 \\

PROB \cite{Zohar_2023_CVPR} &70.4 & 75.4 & 67.3 & 48.1 & 55.9 & 73.5 & 78.5 & 75.4 & 42.8 & 72.2 & \cellcolor{lightgray}  64.2 &\cellcolor{lightgray}  73.8 &\cellcolor{lightgray}  76.0 & \cellcolor{lightgray}  74.8 &\cellcolor{lightgray}  75.3 &\cellcolor{lightgray}  40.2 &\cellcolor{lightgray} 66.2 &\cellcolor{lightgray}  73.3 &\cellcolor{lightgray}  64.4 &\cellcolor{lightgray}  64.0 & 66.5 \\

CAT \cite{cat} & 76.5 & 75.7 & 67.0 & 51.0 & 62.4 & 73.2 & 82.3 & 83.7 & 42.7 & 64.4 & \cellcolor{lightgray}  56.8 &\cellcolor{lightgray}  74.1 &\cellcolor{lightgray}  75.8 &\cellcolor{lightgray}  79.2 &\cellcolor{lightgray}  78.1 &\cellcolor{lightgray}  39.9 &\cellcolor{lightgray}  65.1 &\cellcolor{lightgray}  59.6 &\cellcolor{lightgray}  78.4 &\cellcolor{lightgray}  67.4 & 67.7 \\

OrthogonalDet \cite{sun2024exploring}\footnotemark[1] &82.9 &80.1 &75.8 &64.3 &60.6 &81.5 & 87.9 & 54.9 & 48 &82.1 &\cellcolor{lightgray}  57.7 & \cellcolor{lightgray} 63.5 &\cellcolor{lightgray}  80.5 &\cellcolor{lightgray} 77.6 &\cellcolor{lightgray} 78.2 &\cellcolor{lightgray} 38.9 &\cellcolor{lightgray} 69.8 &\cellcolor{lightgray} 62.8 &\cellcolor{lightgray} 76.9 &\cellcolor{lightgray} 64.2 &69.41 \\

\midrule

\textbf{CROWD (ours)} & 84.1 & 84.5 & 73.9 & 60.0 &	65.1 & 80.1 & 89.3 & 82.7 &	53.3 & 77.4 & \cellcolor{lightgray}63.4 & \cellcolor{lightgray}78.5 &	\cellcolor{lightgray}80.9 & \cellcolor{lightgray}83.4 & \cellcolor{lightgray}83.9 & \cellcolor{lightgray}46.5 & \cellcolor{lightgray}72.6 & \cellcolor{lightgray}60.9 & \cellcolor{lightgray}77.9 & \cellcolor{lightgray}71.5 &	\textbf{73.5} \\

\midrule\midrule

{\textbf{15 + 5 setting}} & aero & cycle & bird & boat & bottle & bus & car & cat & chair & cow & table & dog & horse & bike & person & plant & sheep & sofa & train & tv & mAP \\ \midrule
ILOD \cite{ILOD} & 70.5 & 79.2 & 68.8 & 59.1 & 53.2 & 75.4  & 79.4 & 78.8 & 46.6 & 59.4 & 59.0 & 75.8 & 71.8 & 78.6 & 69.6 & \cellcolor{lightgray}  33.7 & \cellcolor{lightgray}  61.5 & \cellcolor{lightgray}  63.1 &\cellcolor{lightgray}  71.7 &\cellcolor{lightgray}  62.2 & 65.8 \\

Faster ILOD \cite{F_ILOD} & 66.5 & 78.1 & 71.8 & 54.6 & 61.4 & 68.4 & 82.6 & 82.7 & 52.1 & 74.3 & 63.1 & 78.6 & 80.5 & 78.4 & 80.4 & \cellcolor{lightgray} 36.7 &\cellcolor{lightgray}  61.7 &\cellcolor{lightgray}  59.3 &\cellcolor{lightgray}  67.9 &\cellcolor{lightgray}  59.1 & 67.9 \\
ORE ~\cite{joseph2021ore} & 75.4 & 81.0 & 67.1 & 51.9 & 55.7 & 77.2 & 85.6 & 81.7 & 46.1 & 76.2 & 55.4 & 76.7 & 86.2 & 78.5 & 82.1 &\cellcolor{lightgray}  32.8 & \cellcolor{lightgray}  63.6 &\cellcolor{lightgray}  54.7  &\cellcolor{lightgray} 77.7 &\cellcolor{lightgray}  64.6 & 68.5 \\
Meta-ILOD \cite{Meta_ILOD} & 78.4 & 79.7 & 66.9 & 54.8 & 56.2 & 77.7 & 84.6 & 79.1 & 47.7 & 75.0 & 61.8 & 74.7 & 81.6 & 77.5 & 80.2 &\cellcolor{lightgray}  37.8 &\cellcolor{lightgray}  58.0 &\cellcolor{lightgray}  54.6 &\cellcolor{lightgray}  73.0 &\cellcolor{lightgray}  56.1 & 67.8 \\
ROSETTA \cite{ROSETTA} & 76.5 & 77.5 & 65.1 & 56.0 & 60.0 & 78.3 & 85.5 & 78.7 & 49.5 & 68.2 & 67.4 & 71.2 & 83.9 & 75.7 & 82.0 &\cellcolor{lightgray}   43.0 &\cellcolor{lightgray}  60.6 &\cellcolor{lightgray}  64.1 &\cellcolor{lightgray}  72.8 &\cellcolor{lightgray}  67.4 & 69.2 \\
OW-DETR \cite{gupta2021ow}& 77.1 & 76.5 & 69.2 & 51.3 & 61.3 & 79.8 & 84.2 & 81.0 & 49.7 & 79.6 & 58.1 & 79.0 & 83.1 & 67.8 & 85.4 & \cellcolor{lightgray} 33.2 & \cellcolor{lightgray} 65.1 & \cellcolor{lightgray} 62.0 & \cellcolor{lightgray} 73.9 & \cellcolor{lightgray} 65.0 & 69.4 \\ 
PROB \cite{Zohar_2023_CVPR} & 77.9 & 77.0 & 77.5 & 56.7 & 63.9 & 75.0 & 85.5 & 82.3 & 50.0 & 78.5 & 63.1 & 75.8 & 80.0 & 78.3 & 77.2 &\cellcolor{lightgray}  38.4 &\cellcolor{lightgray}  69.8 &\cellcolor{lightgray}  57.1 &\cellcolor{lightgray}  73.7 &\cellcolor{lightgray}  64.9 & 70.1 \\
CAT \cite{cat} &75.3  &81.0 & 84.4 & 64.5 & 56.6  &74.4 & 84.1 & 86.6 & 53.0 & 70.1 & 72.4 & 83.4 & 85.5 & 81.6 & 81.0 &\cellcolor{lightgray} 32.0 &\cellcolor{lightgray}  58.6 &\cellcolor{lightgray}  60.7 &\cellcolor{lightgray}  81.6 &\cellcolor{lightgray}  63.5 & 72.2 \\
OrthogonalDet \cite{sun2024exploring}\footnotemark[1] & 81.8 & 79.3 & 71.0 & 71.0 & 58.8 & 62.1 & 82.6 & 89.7 & 79.8 & 47.0 & 80.5 & 61.1 & 79.9 & 80.2 & 81.6 & \cellcolor{lightgray}44.2 &\cellcolor{lightgray}65.5 &\cellcolor{lightgray}71.5 &\cellcolor{lightgray}75.6 &\cellcolor{lightgray} 74.2 & 72.6 \\
\midrule

\textbf{CROWD (ours)} &  82.8 & 80.6 & 72.5 &59.6 &61.3 &83.1 &89.3 &83 &49.2 &86.1 &62.2 &83.7 &86 &80.3 &82.8 & \cellcolor{lightgray} 46.1 &\cellcolor{lightgray} 80 &\cellcolor{lightgray} 63.7 &\cellcolor{lightgray} 79.5 &\cellcolor{lightgray} 75.6 & \textbf{74.4} \\ 
\midrule\midrule

{\textbf{19 + 1 setting}} & aero & cycle & bird & boat & bottle & bus & car & cat & chair & cow & table & dog & horse & bike & person & plant & sheep & sofa & train & tv & mAP \\ \midrule
ILOD \cite{ILOD} & 69.4 & 79.3 & 69.5 & 57.4 & 45.4 & 78.4 & 79.1 & 80.5 & 45.7 & 76.3 & 64.8 & 77.2 & 80.8 & 77.5 & 70.1 & 42.3 & 67.5 & 64.4 & 76.7 & \cellcolor{lightgray} 62.7 & 68.2 \\

Faster ILOD \cite{F_ILOD} & 64.2 & 74.7 & 73.2 & 55.5 & 53.7 & 70.8 & 82.9 & 82.6 & 51.6 & 79.7 & 58.7 & 78.8 & 81.8 & 75.3 & 77.4 & 43.1 & 73.8 & 61.7 & 69.8 & \cellcolor{lightgray} 61.1 & 68.5 \\

ORE ~\cite{joseph2021ore} & 67.3 & 76.8 & 60 & 48.4 & 58.8 & 81.1 & 86.5 & 75.8 & 41.5 & 79.6 & 54.6 & 72.8 & 85.9 & 81.7 & 82.4 & 44.8 & 75.8 & 68.2 & 75.7 & \cellcolor{lightgray} 60.1 & 68.8 \\ 

Meta-ILOD \cite{Meta_ILOD} &78.2 & 77.5 & 69.4 & 55.0 & 56.0 & 78.4 & 84.2 & 79.2 & 46.6 & 79.0 & 63.2 & 78.5 & 82.7 & 79.1 & 79.9 & 44.1 & 73.2 & 66.3 & 76.4 &\cellcolor{lightgray}  57.6 & 70.2 \\

ROSETTA \cite{ROSETTA} & 75.3 & 77.9 & 65.3 & 56.2 & 55.3 & 79.6 &84.6 &72.9 &49.2 &73.7 &68.3 &71.0 &78.9 &77.7 &80.7 &44.0& 69.6 &68.5& 76.1& \cellcolor{lightgray} 68.3 &69.6 \\

OW-DETR \cite{gupta2021ow} & 70.5 & 77.2 & 73.8 & 54.0 & 55.6 & 79.0 & 80.8 & 80.6 & 43.2 & 80.4 & 53.5 & 77.5 & 89.5 & 82.0 & 74.7 & 43.3 & 71.9 & 66.6 & 79.4 & \cellcolor{lightgray} 62.0 & 70.2 \\

PROB \cite{Zohar_2023_CVPR} & 80.3 &78.9 &77.6 &59.7 &63.7 &75.2 &86.0 &83.9 &53.7 &82.8 &66.5 &82.7 &80.6 &83.8 &77.9 &48.9 &74.5 &69.9 &77.6  &\cellcolor{lightgray}  48.5 &   72.6 \\

CAT \cite{cat} &86.0 &85.8 &78.8 &65.3 &61.3 &71.4 &84.8 &84.8 &52.9 &78.4 &71.6 &82.7 &83.8 &81.2 &80.7 &43.7 &75.9 &58.5 &85.2 &\cellcolor{lightgray} 61.1 & 73.8 \\
OrthogonalDet \cite{sun2024exploring}\footnotemark[1] & 81.8 & 82.6 & 77.0 & 56.3 & 66.0 & 74.4 & 88.5 & 78.7 & 51.2 & 84.3 & 63.1 & 84.4 & 81.3 & 78.8 & 80.9 & 46.8 & 77.9 & 68.6 & 74.1 &\cellcolor{lightgray} 74.5 & 73.6 \\
\midrule
\textbf{CROWD (ours)} &  81.7 &	80.3 & 77.4 & 57.2 & 66.8 & 80.7 &	87.1 & 67.9 & 49.4 & 87.3 & 65.6 & 84.2 & 85.4 & 79.9 & 81.6 & 48.6 & 77.0 & 69.0 & 82.2 & \cellcolor{lightgray}75.3 & \textbf{74.2} \\ 

\bottomrule
\end{tabular}%
}

\end{table*}

\footnotetext[1]{This is a reproduction of the results from OrthogonalDet from their public repo.}

\subsection{Additional Experimental Details}
\label{app:additional_expts}
In this section we provide additional experimental details for training our CROWD approach on M-OWOD, S-OWOD and IOD benchmarks discussed in \cref{sec:expts} of the main paper.

\noindent \textbf{M-OWOD and S-OWOD benchmarks} - M-OWOD and S-OWOD benchmarks are created from MS-COCO~\cite{coco} and split into 4 tasks $T_t$, where $t \in [1,4]$ detailed in the "Datasets" section in \cref{sec:expts}. 
For each task, the model in provided labeled examples from $T_t$ alone while at inference the model is expected to identify objects in tasks leading up to $T_t$, s.t $t \in [1, t]$. We split the training into two splits. In the first stage the model is exposed only to the currently known classes $K^t$ and the learnt model $h^t$ biases on labeled examples in $K^t$. At the end of the first stage CROWD-D kicks in and selects representative unknowns as described in \cref{sec:crowd_d}. Lets call in $U^t$. Next, we store a replay buffer of the currently known objects $\hat{K}^t$, s.t. $\hat{K}^t \subseteq K^t$. Following this, we combine $\hat{K}^t$, $K^t$ and a replay buffer from the previous task $\hat{K}^{t-1}$ into a single dataset to finetune $h^t$ using CROWD-L. As detailed in \cref{sec:crowd_l} this ensures known vs. unknown separation wile retaining discriminative features from known classes.

\noindent \textbf{IOD benchmarks} - In contrast to OWOD, IOD does not encounter unknowns during model training but experiences heavy catastrophic forgetting on previously known classes $K^{t-1}$. Following recent benchmarks like \citet{sun2024exploring, Zohar_2023_CVPR, joseph2021ore} we evaluate the IOD performance of CROWD on PASCAL-VOC benchmark on three settings produced by varying the number of newly added classes - 10 + 10, 15 + 5, 19 + 1 as shown in \cref{tab:iod_benchmark}. In the absence of unknowns we do not apply CROWD-D and only rely on CROWD-L applied to the finetuning stage of IOD. 
Following latest works we adopt a replay based learning technique whichh stores a small subset of the previously known objects $\hat{K}^{t-1}$ in a buffer. $\hat{K}^{t-1}$ combined with the newly introduced classes $K^t$ is used to finetune $h^t$. 
This also requires us to slightly modify the formulation of $L_{\text{CROWD}}^{cross}$ as detailed in \cref{sec:crowd_l}. 
For each setting $h^t$ is trained on a batch size of 12 for 3000 iterations using an AdamW optimizer, a base learning rate to $2.5 \times 10^{-5}$ and weight decay of $1 \times 10^{-4}$.

\subsubsection{Ablation on Selection Budget $\mathtt{k}$}
\label{app:ablation_budget}
As detailed in \cref{sec:crowd_d}, the parameter $\mathtt{k}$ dictates how many candidate unknown RoIs CROWD-D selects per image. We conduct an ablation over several plausible settings of $\mathtt{k}$ within the interval $[0,100]$, and present the outcome in \cref{tab:ablation_k}. Fo this experiment we keep the choice of submodular function $f$ in CROWD-D as Graph-Cut and Facility-Location (CROWD-FL) for CROWD-L following the results of the ablation experiments in \cref{tab:owod_ablations} in the main paper. Notably, raising $\mathtt{k}$ from 0 (i.e., OrthogonalDet) to 10 yields a marked uplift in the model’s unknown–recall (U-Recall), yet further increases beyond $\mathtt{k}=20$ confer no additional gains. This initial boost in U-Recall stems from the integration of truly informative RoIs into the training loop. However, when $\mathtt{k}$ reaches its upper bound of 100, the mean average precision (mAP) on known classes experiences a slight decline relative to existing baselines—a consequence of inadvertently incorporating spurious background proposals.

\begin{table*}[t]
  \centering
  \caption{\textbf{Ablation Experiments on the variation in $\mathtt{k}$ in CROWD-D.} Given the Graph-Cut based selection strategy in \cref{alg:crowd_discover} of CROWD-D and the CROWD-FL based learning objective in CROWD-L we vary $\mathtt{k}$ in [0, 100] and adopt the best performing budget for our pipeline.}
  \resizebox{\textwidth}{!}{%
  \begin{tabular}{l|c|cc|cc|cc|c}
    \toprule
    \textbf{Method} & Budget &
    \multicolumn{2}{c|}{\textbf{Task 1}} & 
    \multicolumn{2}{c|}{\textbf{Task 2}} & 
    \multicolumn{2}{c|}{\textbf{Task 3}} & 
    \textbf{Task 4} \\ 
    
    & $\mathtt{k}$ & 
    \textbf{U-Recall} & \textbf{mAP} & 
    \textbf{U-Recall} & \textbf{mAP} & 
    \textbf{U-Recall} & \textbf{mAP} & 
    \textbf{mAP}  \\
    \midrule \midrule

    OrthogonalDet \cite{sun2024exploring} & - & $24.6$ & $61.3$ & $26.3$ & $47.0$ & $29.1$ & $41.3$ & $37.9$ \\ \midrule
    
    \multirow{4}{*}{\textbf{CROWD (ours)}}   & 5 & 51.2 & 61.0 & 49.1 & 45.9 & 62.7 & 40.3  & 37.4 \\
     & \textbf{10}  & \cellcolor{LightGreen}$57.9$ & \cellcolor{LightGreen}$61.7$ & \cellcolor{LightGreen}$53.6$ & \cellcolor{LightGreen}$47.8$ & \cellcolor{LightGreen}$69.6$ & \cellcolor{LightGreen}$42.4$ & \cellcolor{LightGreen}$38.5$ \\
     & 30  & 58.4 & 61.7 & 53.5 & 48.0 & 70.1 & 42.4 & 38.5  \\ 
     & 100  & 57.5 & 59.3 & 53.7 & 44.3 & 70.9 & 38.8 & 32.0 \\
    \bottomrule
  \end{tabular}}
  \label{tab:ablation_k}
\end{table*}

\subsubsection{Results on Synthetic Datasets - CROWD-D}
\begin{figure*}
        \centering
        \begin{tabular}{rc}
            \rotatebox{90}{\small GCCG} &
                \includegraphics[width=0.98\textwidth]{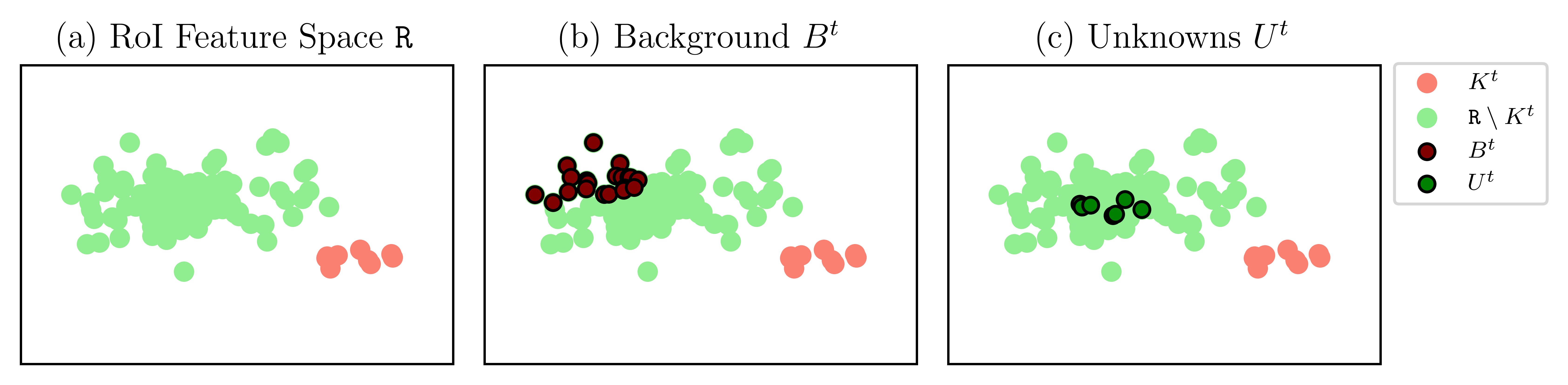} \\
            \rotatebox{90}{\small FLCG } & 
                \includegraphics[width=0.98\textwidth]{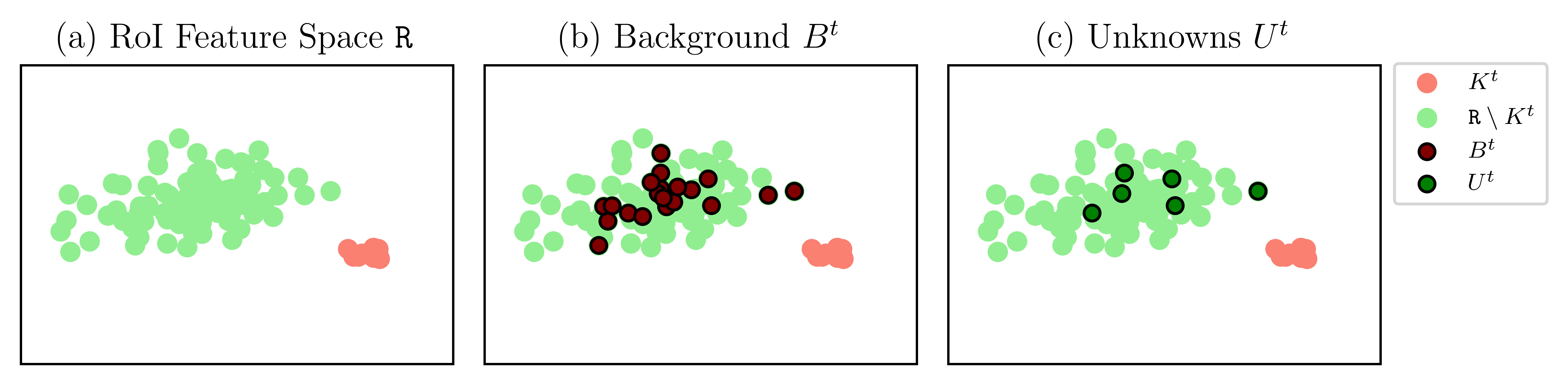} \\
            \rotatebox{90}{\small LogDetCG } & 
                \includegraphics[width=0.98\textwidth]{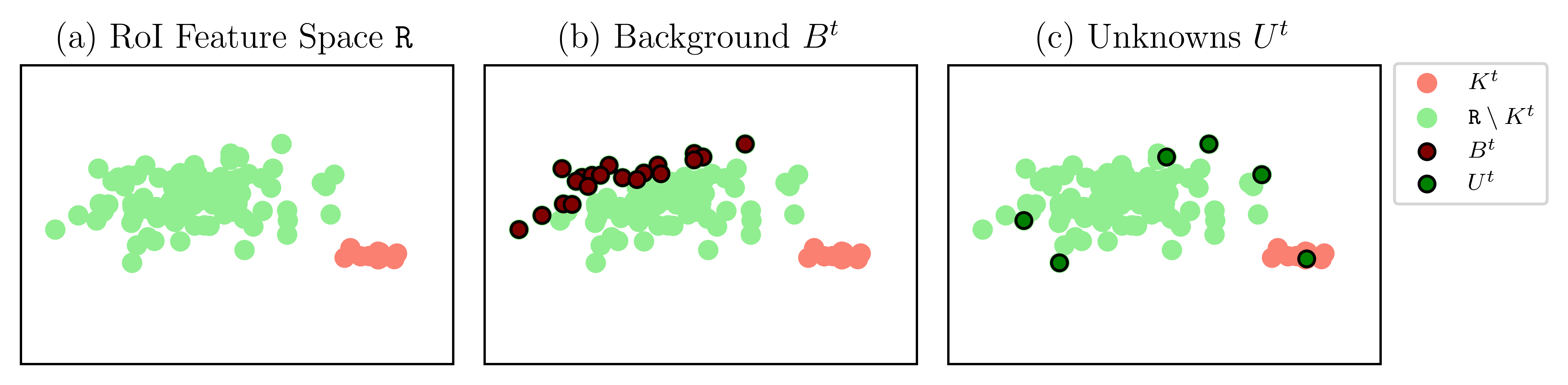} \\
        \end{tabular}
        \caption{\textbf{CROWD-D results on synthetic dataset} contrasted against instances of popular submodular functions - Graph-Cut, Facility-Location and Log-Determinant. Graph-Cut based selection strategy models both representation and diversity resulting in the best possible choice of unknown instances in $U^t$.}
        \label{fig:crowd_d_synth_full}
\end{figure*}
In addition to the illustrations provided in \cref{fig:crowd_selection_gccg} we contrast the selection performance of CROWD-D by varying the underlying submodular function $f$ between Grap-Cut (GC), Facility-Location (FL) and Log-Determinant (LogDet) on synthetic datasets as shown in \cref{fig:crowd_d_synth_full}. 
The use of synthetic datasets provide us with complete control over the embedding space allowing us to pathologically inject imbalance, inter-cluster separation etc. in a compute efficient fashion.
Particularly in our experiments we use a two-cluster imbalanced setup mimicking the RoI embedding space in Faster-RCNN~\cite{frcnn} model. 
Similar to \citet{sun2024exploring} the number of known class and unknown class feature vectors are severely imbalanced with total number of RoIs $\mathtt{R} = 500$ and the number of knowns $|K^t| = 10$. $\mathtt{R}$ and $K^t$ are sampled from a normal distribution with fixed variance values. 
The LogDet based selection strategy enforces the notion of diversity in the selection mechanism which does not select representative unknowns negatively impacting OWOD performance as shown in \cref{tab:owod_ablations}. 
The FL based selection strategy models representation as shown in \cref{fig:crowd_d_synth_full} alone during selection resulting in erroneous selection of background instances negatively affecting OWOD performance.
Lastly, GC based selection strategy shown in \cref{fig:crowd_d_synth_full} models notions of both diversity and representation selecting diverse backgrounds $B^t$ farthest to $K^t$ as well as representative unknowns $U^t$. This results in GC based selection strategy to produce the best overall results as shown in \cref{tab:open_world_results}.



\end{document}